\documentclass[lettersize,journal]{IEEEtran}
\usepackage{amsmath,amsfonts}
\usepackage{algorithmic}
\usepackage{algorithm}
\usepackage{array}
\usepackage[caption=false,font=normalsize,labelfont=sf,textfont=sf]{subfig}
\usepackage{textcomp}
\usepackage{stfloats}
\usepackage{url}
\usepackage{verbatim}
\usepackage{graphicx}
\usepackage{cite}
\usepackage{color}
\usepackage{bbm}
\usepackage{multirow}
\usepackage[normalem]{ulem}
\begin{document}

\title{Learning Pixel-wise Continuous Depth Representation via Clustering for Depth Completion}

\author{Shenglun Chen$^1$, Hong Zhang$^2$, Xinzhu Ma$^3$, Zhihui Wang$^{4*}$,~\IEEEmembership{Member,~IEEE}, Haojie Li$^2$

\thanks{This work is supported in part by the National Natural Science Foundation of China (NSFC) under Grants No.61976038, NO. 61932020 and NO. U1908210, and the Taishan Scholar Program of Shandong Province.}

\thanks{$^{1}$S. Chen is with Dalian University of Technology, Dalian 116620, China. {\tt\small shenglunch@gmail.com}.}

\thanks{$^{2}$H. Zhang and H. Li are with the College of Computer Science and Engineering, Shandong University of Science and Technology, Qingdao 266590, China. {\tt\small hongzh726@gmail.com, hjli@sdust.edu.cn}.}

\thanks{$^{3}$X. Ma is with Shanghai Artificial Intelligence Laboratory / Shanghai AI Lab, Shanghai 200030, China. {\tt\small maxinzhu@pjlab.org.cn}.}

\thanks{$^{4}$Z. Wang is with the DUT-RU International School of Information Science and Engineering, Dalian University of Technology, Dalian 116620, China, and also with the Key Laboratory for Ubiquitous Network and Service Software of Liaoning Province, Dalian University of Technology, Dalian 116620, China. Z. Wang is the corresponding author. {\tt\small zhwang@dlut.edu.cn}.}
}

\markboth{Journal of \LaTeX\ Class Files,~Vol.~14, No.~8, August~2021}%
{Shell \MakeLowercase{\textit{et al.}}: A Sample Article Using IEEEtran.cls for IEEE Journals}


\maketitle

\begin{abstract}
Depth completion is a long-standing challenge in computer vision, where classification-based methods have made tremendous progress in recent years. However, most existing classification-based methods rely on pre-defined pixel-shared and discrete depth values as depth categories. This representation fails to capture the continuous depth values that conform to the real depth distribution, leading to depth smearing in boundary regions. To address this issue, we revisit depth completion from the clustering perspective and propose a novel clustering-based framework called CluDe which focuses on learning the pixel-wise and continuous depth representation. The key idea of CluDe is to iteratively update the pixel-shared and discrete depth representation to its corresponding pixel-wise and continuous counterpart, driven by the real depth distribution. Specifically, CluDe first utilizes depth value clustering to learn a set of depth centers as the depth representation. While these depth centers are pixel-shared and discrete, they are more in line with the real depth distribution compared to pre-defined depth categories. Then, CluDe estimates offsets for these depth centers, enabling their dynamic adjustment along the depth axis of the depth distribution to generate the pixel-wise and continuous depth representation. Extensive experiments demonstrate that CluDe successfully reduces depth smearing around object boundaries by utilizing pixel-wise and continuous depth representation. Furthermore, CluDe achieves state-of-the-art performance on the VOID datasets and outperforms classification-based methods on the KITTI dataset.
\end{abstract}


\section{Introduction}
\IEEEPARstart{G}EOMETRY information plays a crucial role in various applications and technologies related to environment perception, including robotics, autonomous navigation, augmented reality, and virtual reality. To capture accurate geometry information, active ranging sensors like LiDAR, radar, and Kinect have been extensively employed in these applications to measure the depth of the environment. However, inherent limitations of these sensors generally result in depth maps with significant holes, which hampers the effectiveness of environmental perception technology. To overcome this challenge, numerous depth completion methods \cite{ma2018sparse,liu2021fcfr,10173567,10243117,10064324} have been proposed to generate a complete depth map by filling in the sparse depth map using RGB image guidance.

\begin{figure}[h]
\centering
\includegraphics[width=0.9\linewidth]{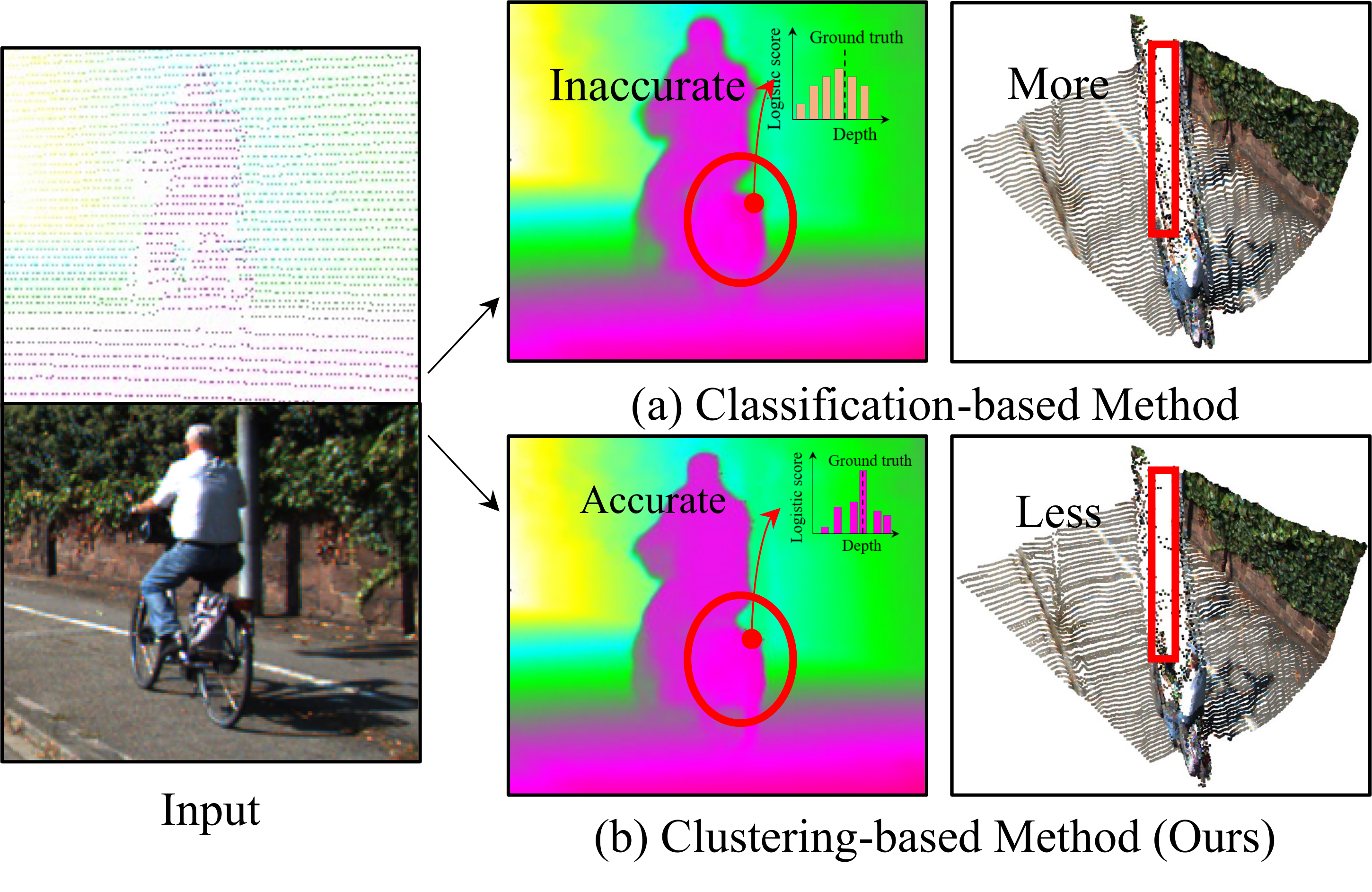}
\caption{Results of classification-based and clustering-based methods. It is observed that the clustering-based method produces more accurate logistic scores, thereby preserving the shape of the object while effectively suppressing stripe artifacts between the person and the wall.}
\label{fig_mo}
\end{figure}

\begin{figure*}[t]
\centering
\includegraphics[width=0.9\linewidth]{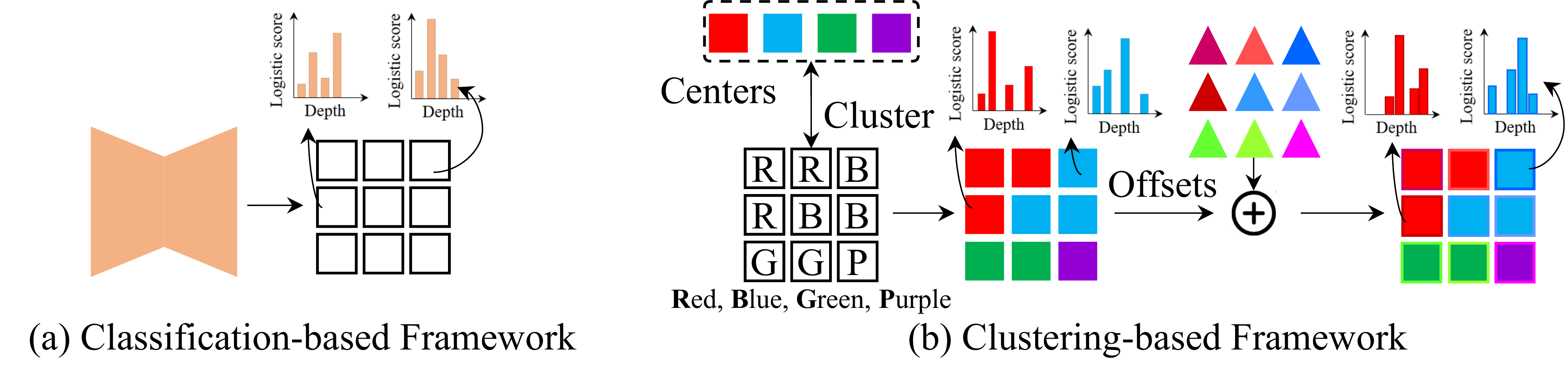}
\caption{In the classification-based framework, the main purpose is to predict logistic scores for pre-defined depth categories. In contrast, the proposed clustering-based framework focuses on obtaining clusters via depth centers. Additionally, we estimate offsets to improve the continuity of depth for each pixel.} 
\label{fig_pip}
\end{figure*}
The depth completion is commonly approached as a regression-based problem, similar to the RGB-based depth estimation \cite{10.1145/3503161.3548381}. By leveraging convolutional neural networks (CNNs) trained with the regression loss function, depth completion methods are able to infer complete depth information from none \cite{lee2021depth}. While these regression-based methods show encouraging progress, they still suffer from slow convergence and depth smearing between objects \cite{fu2018deep,li2022binsformer,imran2019depth}. 

As a result, other depth completion methods explore the classification-based framework to solve these issues \cite{imran2019depth,lee2021depth,kam2022costdcnet,peng2022pixelwise}. Similar to semantic segmentation \cite{long2015fully}, these classification-based methods first pre-define a set of depth values or planes as potential depth categories based on depth distribution. The depth distribution denotes the probability distribution of depth values within a given continuous depth range for a specific image or dataset, while depth categories serve as the representation (discretization) of depth according to the depth distribution. Then, these methods adopt the cross-entropy loss function to train the network to predict logistic scores for these depth categories. Instead of selecting the most significant category as the final depth result, depth completion accumulates all logistic scores to calculate the mathematical expectation, which theoretically allows for arbitrary depth result within the continuous depth range. 

Therefore, most classification-based methods focus on training a classifier to predict logistic scores over pre-defined depth categories. However, two main challenges are neglected: \textit{(1)} The real depth distribution is not a simple uniform distribution and varies across different scenes in different images, making it challenging to directly determine depth categories representing the depth values under various depth distributions. \textit{(2)} Pixel-shared and discrete depth categories fail to describe the continuous depth values and cannot capture diverse depth distributions of pixels caused by distinct contexts and structures. These problems give rise to ambiguous logistic scores, leading to an inaccurate depth result. As illustrated in Fig. \ref{fig_mo} (a), artifacts are observed in the point cloud, such as streaking between the person and the wall.

To address the challenges mentioned above, it is crucial to represent the continuous depth values observed in real-world scenes with various depth distributions. Drawing inspiration from case-based reasoning in visual recognition \cite{kolodner1992introduction,wang2022visual}, we propose a clustering-based framework for depth completion named CluDe, which focuses on learning pixel-wise and continuous depth representation from real-world scenes. As shown in Fig. \ref{fig_mo} (b), our method successfully suppresses stripe artifacts between objects.

In contrast to the typical classification-based framework, our proposed clustering-based framework consists of two main stages, as illustrated in Fig. \ref{fig_pip}. In the first stage, we introduce a clustering approach that focuses on learning a set of feature vectors as depth centers to represent the real depth distribution of the dataset. In the second stage, we adopt offset estimation to convert these centers into pixel-wise and continuous representations, which are specific to the current input data.

Specifically, in the first stage, we propose a transformer structure dedicated to learning depth centers from real-world scenes to support clustering. Depth centers are defined as a set of learnable feature vectors intended to capture the depth distribution across the entire dataset. Throughout the clustering process, these depth centers act as the initial depth centers. Then, we exploit the global self-attention mechanism within the transformer to cluster the depth information of current input based on these initial depth centers. As the clustering process progresses, the initial depth centers gradually align with the depth distribution of the current input, resulting in current depth centers.

Learning real depth centers is challenging due to the absence of direct supervision. To address this issue, we adopt an implicit approach to learn depth centers. We introduce depth guidance as a hint for the depth centers and jointly estimate the depth result using these centers. By supervising the depth result, we implicitly learn initial depth centers.

Nevertheless, the depth centers obtained from the first stage are limited and pixel-shared for current input. This forces CluDe to predict depth result through a single discrete depth representation, leading to depth smearing. Therefore, in the second stage, we introduce a hierarchical translation module which can gradually estimate multiple depth offsets for refinement. These depth offsets allow us to dynamically adjust the depth guidance, translating the depth representation along the depth axis of the distribution. This transformation converts the single discrete depth representation into a pixel-wise and continuous depth representation.

We evaluate the performance of CluDe on both outdoor and indoor scenes (KITTI \cite{geiger2012we} and VOID \cite{wong2020unsupervised} datasets). Qualitatively, CluDe generates dense depth maps efficiently while preserving clear boundaries between objects. Quantitatively, our method achieves state-of-the-art performance in classification-based depth completion methods and yields competitive results in the boundary region when compared to typical methods.

To summarize, our main contributions are listed as follows:
\begin{itemize}
\item To the best of our knowledge, we are the first to regard the depth completion from a clustering perspective, and propose a clustering-based framework called CluDe.
\item Our proposed CluDe can learn pixel-wise and continuous depth representation. Specifically, CluDe first implicitly learns a set of depth centers from real depth distribution as the discrete depth representation, and then dynamically translates it along the depth axis to convert pixel-wise continuous depth representation.
\item Our method yields competitive results in the boundary region, and achieves state-of-the-art performance among classification-based depth completion methods on KITTI and VOID dataset.
\end{itemize}

\section{Related works}
\subsection{Depth Completion}
The aim of depth completion is to generate a complete depth map from a sparse depth map using the corresponding RGB image guidance. Existing methods in this task can be categorized into two groups, depending on the approach employed to estimate the depth result: regression-based and classification-based methods.

Most regression-based depth completion methods primarily focus on efficient feature fusion and enhancing spatial constraints. Inspired by the guided image filtering, GuideNet \cite{9286883} learn kernel weights from the guidance image, containing content-dependent and spatially-variant kernels to extract the depth image features for multi-modal feature fusion. DGDF \cite{10173567} decomposes the guided dynamic filters into a combination of content-adaptive adaptors and a spatially-shared component to exploit the multimodal information. FCFR-Net \cite{liu2021fcfr} formulates depth completion as a two-stage learning task, which involves a channel shuffle extraction operation to extract representative features from the RGB image and coarse depth map, as well as an energy-based fusion operation to effectively combine these features. GuideFormer \cite{rho2022guideformer} is a fully transformer-based architecture that processes the sparse depth map and RGB image with separate branches. It utilizes guided-attention to capture inter-modal dependencies, enabling it to handle diverse visual dependencies and recover depth values while preserving fine details. DeepLiDAR \cite{Qiu_2019_CVPR} predicts a confidence mask to handle mixed LiDAR signals near foreground boundaries. It combines estimates from the color image and surface normal using learned attention maps to improve depth accuracy, especially for distant areas. ACMNet \cite{zhao2021adaptive} replaces standard convolution operation with graph propagation to capture spatial contexts. It constructs multiple scaled graphs using observed pixels and applies an attention mechanism to model contextual information adaptively.

Additionally, some regression-based methods employ the spatial propagation network (SPN) \cite{liu2017learning} to refine the depth result rather than aiming for a perfect regression result directly. CSPN \cite{cheng2018depth} is the first to introduce SPNs into depth completion. The method utilizes convolution kernels to directly extract directional information and iteratively refines the initial depth estimation using affinities. Building upon the success of CSPN, subsequent works such as NLSPN \cite{park2020non}, DySPN \cite{lin2022dynamic,10284921}, and GraphCSPN \cite{liu2022graphcspn} have further improved the performance of SPNs for depth completion. 

Nonetheless, direct regression from none to obtain depth results often encounters challenges, such as slow convergence and excessive smoothing in boundary regions \cite{fu2018deep,li2022binsformer,imran2019depth}. Therefore, classification-based depth completion methods are proposed to solve the problems and show promising performance. DC \cite{imran2019depth} is a pioneering work that introduces a classification-based framework for depth completion, which proposes depth coefficients as representation of depth values to avoid inter-object depth mixing, and utilizes a cross-entropy loss function to promote the performance. CostDCNet \cite{kam2022costdcnet} proposes a depth completion framework based on the cost volume-based depth estimation method commonly used in multi-view stereo task, and introduces three designs of RGB-D feature volume to represent a single RGB-D image into 3D space, leading to better performance. 

However, the aforementioned classification-based methods use pre-defined depth categories to represent depth values. These depth categories are pixel-shared and discrete failing to capture the continuity of depth values. As a result, this unsatisfactory representation may impose limitations on the quality of depth completion and cause depth smearing, particularly in boundary regions. Lee \textit{et al.} \cite{lee2021depth} address this problem by reformulating depth completion into a combination of depth plane classification and residual regression. The method introduces a plane-residual representation where the depth value is represented by the nearest depth plane and a residual value that denotes the difference between depth plane and real depth value. By incorporating residual regression, the continuity of depth plane classification is enhanced, leading to improved performance. Following the typical classification-based framework, PADNet \cite{peng2022pixelwise} introduces pixel-wise adaptive discretization and uncertainty sampling to adaptively assign depth categories that represent the depth values in local range. The method first estimates depth median using pre-defined depth categories, and then determines the local depth range using uncertainty to assign depth categories again, resulting in pixel-wise adaptive depth categories and better performance. But the depth representation remains discrete.

In this work, we focus on the classification-based depth completion method and place an emphasis on addressing depth smearing between objects. Rethinking classification-based depth completion from a clustering perspective, we propose a clustering-based framework named CluDe. The purpose is to learn a pixel-wise and continuous depth representation that effectively captures the continuous depth values across various real depth distributions.

\subsection{Transformer}
Motivated by the success of transformers in natural language processing \cite{vaswani2017attention}, the vision transformer (ViT) \cite{dosovitskiy2021an} has emerged and demonstrated remarkable achievements in computer vision. The standard ViT approach involves dividing the input image into multiple patches, linearly embedding each patch, adding positional embeddings, and feeding the resulting sequence of vectors into a conventional transformer encoder. The output corresponding to the special token [CLS] is treated as the final output of the encoder, representing the final image representation. ViT is widely used in various computer vision tasks, such as object detection \cite{wang2021pyramid}, semantic segmentation \cite{liu2021swin} and depth completion \cite{rho2022guideformer}. GroupViT \cite{xu2022groupvit} is a outstanding work for zero-shot semantic segmentation. This method leverages a transformer-based architecture to effectively learn semantic representation from natural language via contrast learning, without relying on any pixel-level annotations. GroupViT successfully explores the potential of transformer by fully utilizing the global self-attention mechanism, which can estimate arbitrary image segments, going beyond the limitations of rectangular-shaped segments.

Inspired by GroupViT, we design a clustering transformer specifically for depth completion. Different from the standard transformer, our clustering transformer learns depth centers as the discrete depth representation following real depth distribution, and groups input tokens into depth clusters according to similarities from the self-attention mechanism.

\begin{figure*}[t]
\centering
\includegraphics[width=0.9\linewidth]{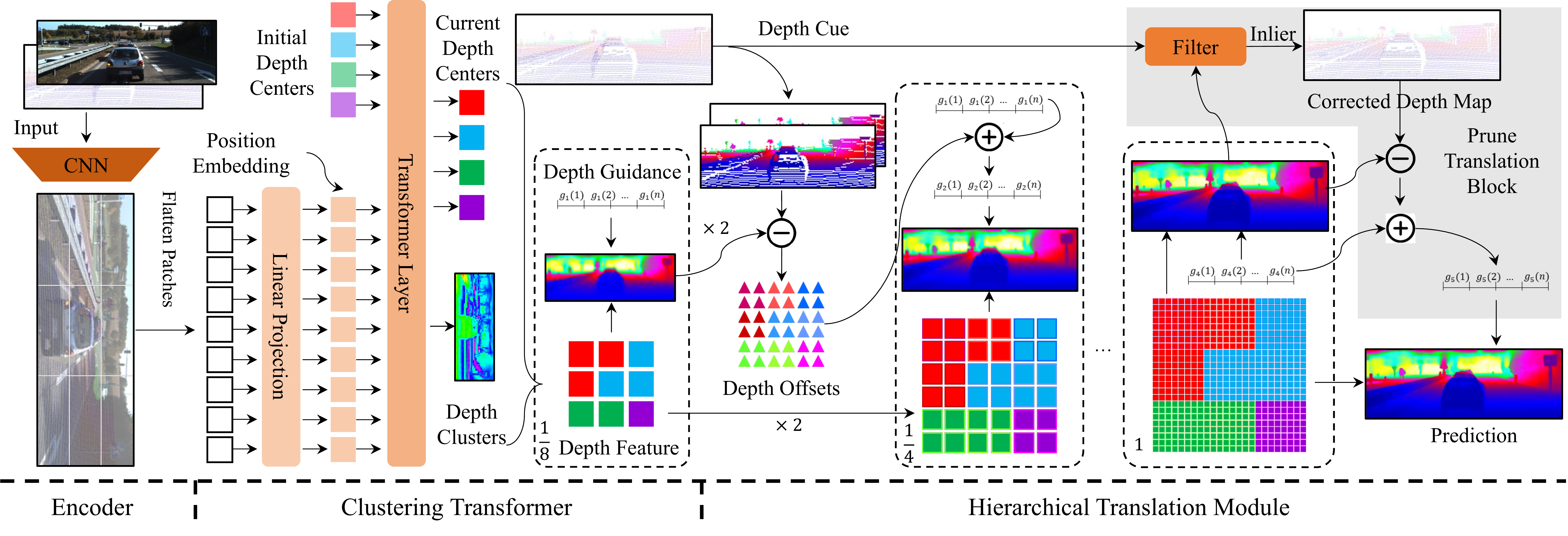}
\caption{The architecture of CluDe. Given a pair of sparse depth map and RGB image, CluDe first extracts multi-scale features. Then, the lowest scale feature is fed into a clustering transformer, which groups tokens into depth clusters and predicts depth result. Next, a hierarchical translation module estimates depth offsets to adjust depth guidance for refinement. For clarity, we omit some features used in depth predication and iteration.}
\label{fig_arch}
\end{figure*}
\subsection{Offset Estimation}
Offset estimation is widely used in computer vision tasks to align features from different scales and modalities. In the pose estimation, offset estimation plays a crucial role in generating exact keypoints and joints \cite{papandreou2017towards}. For instance, A2J \cite{xiong2019a2j} predicts in-plain offsets between the anchor points and joints to obtain 3D position. Similarly, JGR \cite{fang2020jgr} exploits dense pixel-wise offset from pixel to joints to refine their positions, achieving improved pose estimation results. In the stereo matching, CDN \cite{garg2020wasserstein} incorporates offset estimation to address disparity ambiguity. By adjusting the disparity distribution through offset estimation, CDN effectively reduces the uncertainty and improves the performance of stereo matching. In the semantic segmentation, both AlignSeg \cite{huang2021alignseg} and SFANet \cite{weng2021stage} learn pixel-wise offsets based on flow, and then the offsets are used to reconstruct aligned features in different network layers, leading to improved performance in the task.

We introduce offset estimation to estimate the offsets of depth representation. With these offsets, we can continuously convert the learned pixel-shared and discrete depth representation along the depth axis of depth distribution into the pixel-wise and continuous depth representation, and improve the performance accordingly.

\section{METHODOLOGY}
\subsection{Problem Definition}
The goal of depth completion is to generate a complete depth map from a pair of sparse depth map and RGB image. The standard mathematical representation is defined as follows:
\begin{equation}
	D = \Theta(S, I),
\end{equation}
where $D \in \mathbb{R}^{1\times H\times W}$, $S \in \mathbb{R}^{1\times H\times W}$, and $I \in \mathbb{R}^{3\times H\times W}$ represent the complete depth map, sparse depth map, and RGB image, respectively. $\Theta$ denotes a depth completion network. In our proposed clustering-based framework, we define $K$ learnable feature vectors with $M$ dimensions, denoted as initial depth centers $C \in \mathbb{R}^{M\times K}$ in the clustering process. Thus, the problem can be defined as follows:
\begin{equation}
	D = \Theta(S, I, C).
\end{equation}
\subsection{Overview}
We regard the problem of depth completion from a clustering perspective and propose a novel neural network called CluDe that learns pixel-wise and continuous depth representation to address depth smearing, as illustrated in Fig. \ref{fig_arch}. 
Given a pair of sparse depth map and RGB image, we first use an encoder to extract multi-scale feature maps, from $\frac{1}{8}$ to $1$ scales. Then, we introduce a transformer to support such a clustering-based framework, using a set of learnable feature vectors as initial depth centers. In the clustering process, we group the input into non-grid-like depth clusters and update the initial depth centers to yield current depth centers. Based on these produces, we reconstruct a new depth feature to predict the depth result. Finally, we refine the depth result using a hierarchical translation module. This module iteratively estimates depth offsets based on the depth cue to refine the depth result. Subsequent sections will present a detailed introduction to the proposed CluDe.

\subsection{Encoder}
We adopt an encoder network to extract multi-scale features for subsequent clustering and refinement. The input comprises the sparse depth map and RGB image. Additionally, we convert the sparse depth map into the logistic score as an extra input, aiming to enrich the information related to the depth category. We adopt an approximate Laplace distribution to transform the logistic score \cite{chen2019over}, as follows,
\begin{equation}
	l_i = \frac{\exp(-|d-g_i|) }{\sum_{j} \exp(-|d-g_j|)},
\end{equation}
where $d$ is a depth value in the sparse depth map, $g_i$ is the $i$th depth guidance, and $l_i$ is the corresponding logistic score which also denotes the similarity between $d$ and $g_i$. If there is a zero value in the sparse depth map, the corresponding logistic scores are also zero. We use uniform sampling to sample the depth guidance across the entire depth range, which imply the initial depth centers of the dataset. Consequently, the logistic score reflects the similarity between sparse depth and each depth centers, providing initial classification information.  

We exploit three different $3 \times 3$ convolutional layers to pre-process these three input data respectively, and then feed them into the encoder. The encoder follows a unet-like network structure, consisting of residual blocks and spatial pyramid pooling \cite{shen2021cfnet}. The extracted multi-scale features are denoted as $\{ F_i \}_{i=(1,2,3,4)}$, corresponding to scales $\frac{1}{8}$, $\frac{1}{4}$, $\frac{1}{2}$, and the original scale, respectively.

\subsection{Clustering Transformer}
In order to achieve depth clustering, we introduce a transformer structure where global attention mechanism naturally supports our clustering-based framework. In contrast to the standard transformer \cite{dosovitskiy2021an}, which processes all input tokens through all layers, similar tokens are grouped together to form clusters in our clustering transformer.

The input of our clustering transformer is a sequence of feature vectors. To balance the computational cost, we only utilize the $\frac{1}{8}$ scale feature map $F_1$ to yield this sequence. We consider each point in the feature map as a non-overlapping patches and flatten them into a sequence. Then, we linearly project each patch into a latent space and add Fourier position embedding to retain positional information, resulting in $N$ embedding tokens. The subsequent stage consists of \textit{(1)} depth centers-based clustering and \textit{(2)} depth result prediction.

\begin{figure}[h]
\centering
\includegraphics[width=0.9\linewidth]{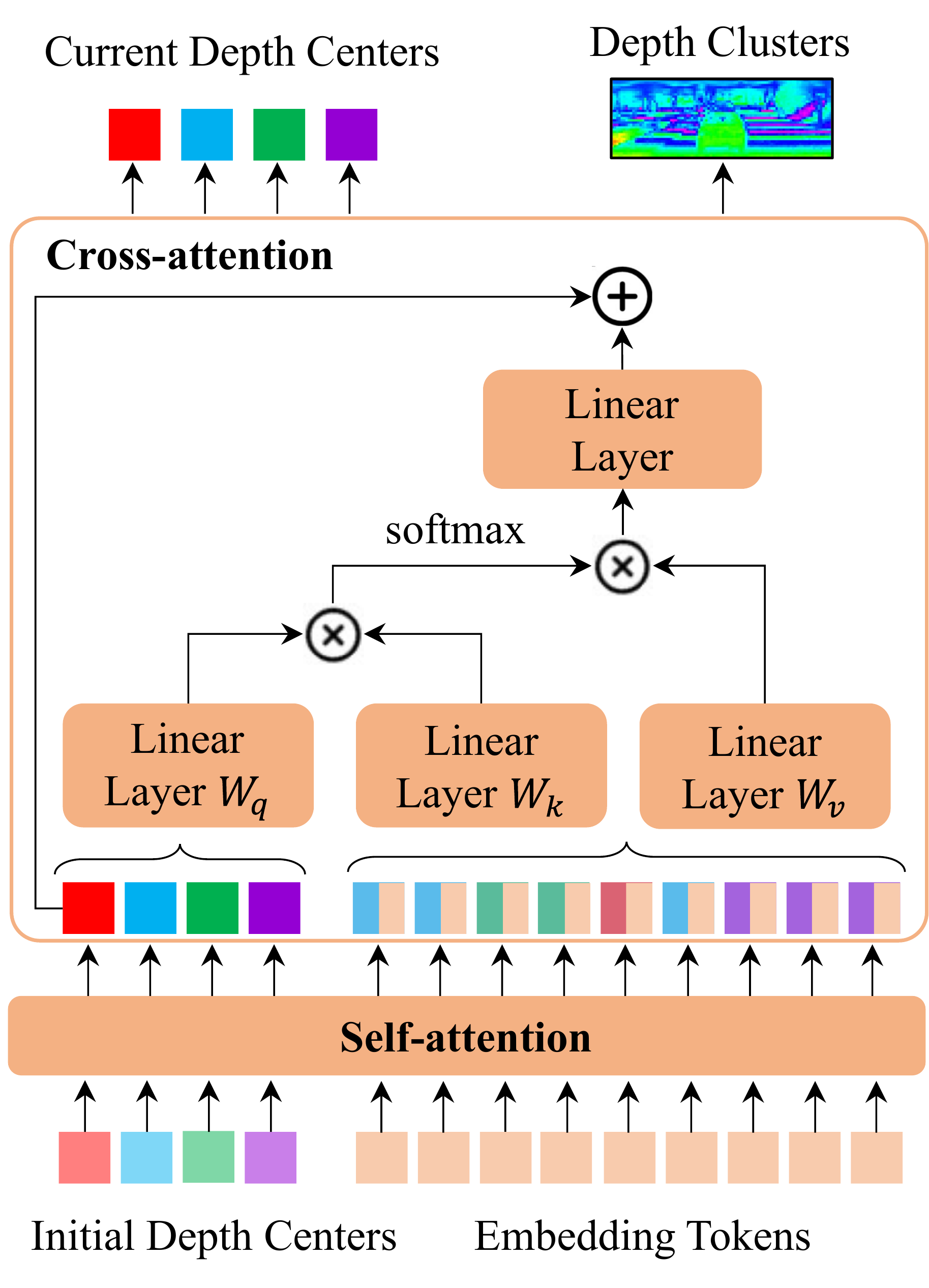}
\caption{The pipeline of clustering. Self-attention is used to propagate information between depth centers and embedding tokens, while cross-attention is utilized to group similar tokens together. This process updates initial depth centers and simultaneously generates depth clusters.}
\label{fig_ca}
\end{figure}

{\setlength{\parindent}{0cm}\textbf{Depth centers-based clustering}.} This stage is the core of the clustering transformer, aiming to aggregate each embedding token into the nearest cluster to obtain depth clusters. As shown in Fig. \ref{fig_ca}, we define a set of learnable feature vectors as initial depth centers $\{c_i\}_{i=1}^{K}$. We feed both these initial depth centers and embedding tokens $\{t_i\}_{i=1}^{N}$ into the transformer layer. The internal attention mechanism of transformer allows depth information to be propagated between depth centers and current input. Simultaneously, we update them to $\{\hat{c}_i\}_{i=1}^{K}$ and $\{\hat{t}_i\}_{i=1}^{N}$. Next, we exploit a cross-attention block \cite{xu2022groupvit} where the attention mechanism naturally provides the ability to calculate the similarities between these two kinds of tokens,
\begin{equation}
	A_{ij} = \frac{\exp(W_q\hat{c}_i \cdot W_k \hat{t}_j) }{\sum_{k=1}^{K} \exp(W_q\hat{c}_k \cdot W_k \hat{t}_j)},
\end{equation}
where $W_q$ and $W_k$ are the weights of the linear projections, and $A  \in \mathbb{R}^{K\times N}$ is the similarity matrix indicating the depth clusters. For simplicity, we omit several necessary transpose operations. Next, we integrate tokens and depth clusters to yield current depth centers,
\begin{equation}
	\tilde{c}_i = \sum_{j=1}^{N} A_{ij} \cdot W_v\hat{t}_j.
\end{equation}
Different from the initial depth centers, which capture the depth distribution across all samples, the current depth centers are adjusted to align with the depth distribution specific to the current input.

With the clustering transformer, CluDe learns a set of high-level representation as the initial depth centers through the statistic of all depth values in the whole dataset, and then updates them to capture the depth distribution of current input. As there is no explicit supervision available for both the initial and current depth centers, CluDe implicitly learns the initial depth centers by supervising the depth result generated by the current depth centers.

{\setlength{\parindent}{0cm}\textbf{Depth result prediction}.} We design this stage to predict the depth result based on the output of the clustering process. Intuitively, we can select the closest depth center based on similarity to represent the depth value of one pixel. However, since the depth value is infinite and the number of $\tilde{c}$ is limited, it is not possible to accurately represent the exact depth value. Additionally, $\tilde{c}$ serves as a feature representation rather than a direct representation of the real depth value. 
To predict depth result, we exploits current depth centers $\{\tilde{c}_i\}_{i=1}^{K}$ and depth clusters $A$ to reconstruct a depth feature $\tilde{F_1}$ with the same shape as $F_1$. This reconstructed feature is then fed into a depth estimation head network to predict depth result,
\begin{equation}
	\begin{split}
		& L_1 = {\rm H}_1(\tilde{F_1}), \\
		& D_1(x, y) = \sum_{i=1}^{K} L_1(x, y, g_1(i)) \cdot g_1(i), 
	\end{split}
\end{equation}
where $\rm{H}_1$ is the depth estimation head network, $L_1(x, y, g_1(i))$ is the $i$th logistic score at the coordinate $(x, y)$ for initial depth guidance $g_1(i)$, and $D_1$ is the depth prediction at $\frac{1}{8}$ scale. The initial depth guidance is a set of depth values uniformly sampled from the given depth range. In fact, the initial depth guidance can be regarded as a hint for the depth centers. By supervising the $D_1$, the clustering transformer focuses on implicitly learning a set of feature vectors as initial depth centers of the dataset.

\subsection{Hierarchical Translation Module}
Our hierarchical translation module is designed to refine the depth result. Although learnable depth centers are effective in describing the real depth distribution, they are shared among pixels for a single input and tend to be discrete, potentially resulting in ambiguous result. Our proposed hierarchical translation module can gradually refine the depth representation to address this problem.

Given the constraint that the number of depth centers cannot increase infinitely, we need to progressively adjusts the depth centers to construct the pixel-wise continuous depth representation for refinement. Specifically, we first estimate the adjustment offset at the $\frac{1}{4}$. To precisely determine the offset, we introduce the depth cue of the initial depth result, which can indicate the direction of adjustment,
\begin{equation}
	C_2 = S_2 - {\rm Up}(D_1),
\end{equation}
where $C_2 \in \mathbb{R}^{2\times \frac{H}{4} \times \frac{W}{4}}$ is the depth cue, $S_2 \in \mathbb{R}^{2\times \frac{H}{4} \times \frac{W}{4}}$ is the sparse depth map at the corresponding scale, and $\rm{Up}$ is the bilinear interpolation for 2 times upsampling.
To preserve more structure of the sparse depth map, we employ both \textit{max-pooling} and \textit{min-pooling} operations to generate sparse depth maps at low scale from the original sparse depth map, and concatenate them together. For the original scale, we directly use the input sparse depth map to calculate the depth cue. To capture both global and local depth variations, we estimate global and local depth offsets for each pixel through the depth cue,
\begin{equation}
	O^g_2, O^l_2 = {\rm O}_2(F_2, C_2),
\end{equation}
where ${\rm O}$ is the offset estimation network at the corresponding scale, which only consists of a residual block, a global offset regressor without any activation function, and a local offset regressor with the \textit{tanh} activation function. $O^g_2 \in \mathbb{R}^{1\times \frac{H}{4} \times \frac{W}{4}}$ represents the global offset that specifically targets global depth variation, and we exploit it to uniformly translate all depth guidance, aiming to improve global continuity of the depth representation. $O^l_2\in \mathbb{R}^{K\times \frac{H}{4} \times \frac{W}{4}}$ represents the local offset that enhances local continuity, which can fine-tune each depth guidance in the narrow band between $g_1(i)$ and $g_1(i+1)$. We can generate the adjusted depth guidance as follows,
\begin{equation}
	g_2(x, y, i) = g_1(i) + O^g_2(x, y) + O^l_2(x, y, i),
\end{equation}
where $g_2(x, y, i)$ is the $i$th adjusted depth guidance at the coordinate $(x, y)$ based on $g_1(i)$. Then, we can predict the corresponding depth result via depth feature $\tilde{F_2} = {\rm Up}(\tilde{F_1})$ and depth guidance $g_2$, 
\begin{equation}
\label{eq_sd2}
\begin{split}
& L_2 = {\rm H}_2(\tilde{F_2}), \\
& D_2(x, y) = \sum_{i=1}^{K} L_2(x, y, g_2(x, y, i)) \cdot g_2(x, y, i),
\end{split}
\end{equation}
where $D_2$ is the depth result at $\frac{1}{4}$ scale. Compared to the pixel-shared initial depth guidance, adjusted depth guidance is pixel-wise which can provide fine-grained hint for current depth centers. As shown in Fig. \ref{fig_arch}, we repeat the above operations to generate depth results $D_3$ at $\frac{1}{2}$ scale and $D_4$ at original scale.

However, a potential problem is that the sparse depth map contains plenty of outliers due to occlusion and calibration errors. These outliers can have an impact on the accuracy of depth cues. For $\frac{1}{4}$ and $\frac{1}{2}$ scales, the \textit{max-pooling} and \textit{min-pooling} operations can be treated as filters that help remove some noise. At the original scale, the noise is carried over to the depth cue. Hence, we introduce a prune translation block to correct the noise and further refine the depth result.

\begin{figure}[h]
\centering
\includegraphics[width=0.9\linewidth]{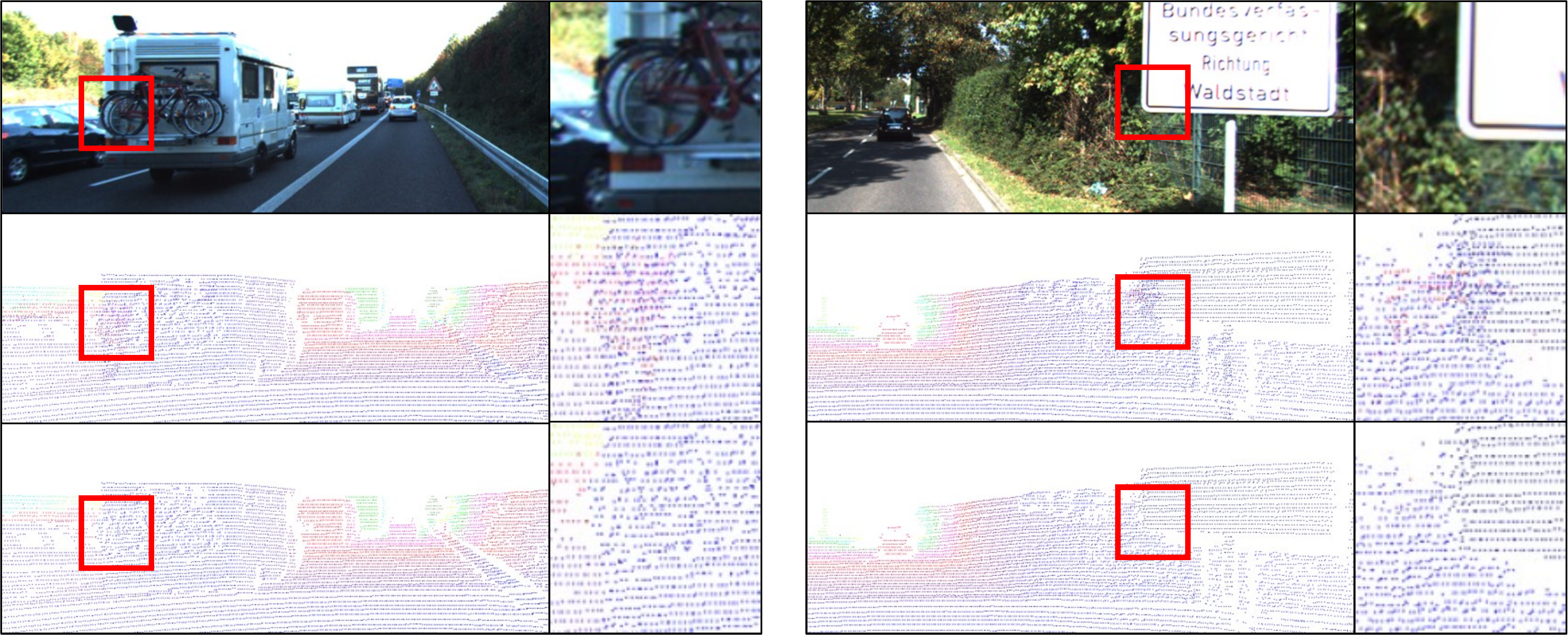}
\caption{From top to bottom, we show the RGB image, sparse depth map, and corrected sparse depth map. The region containing outliers is highlighted with a red box, and a zoomed-in view is provided on the right.}
\label{fig_filter}
\end{figure}

\begin{figure*}[t]
\centering
\includegraphics[width=1\linewidth]{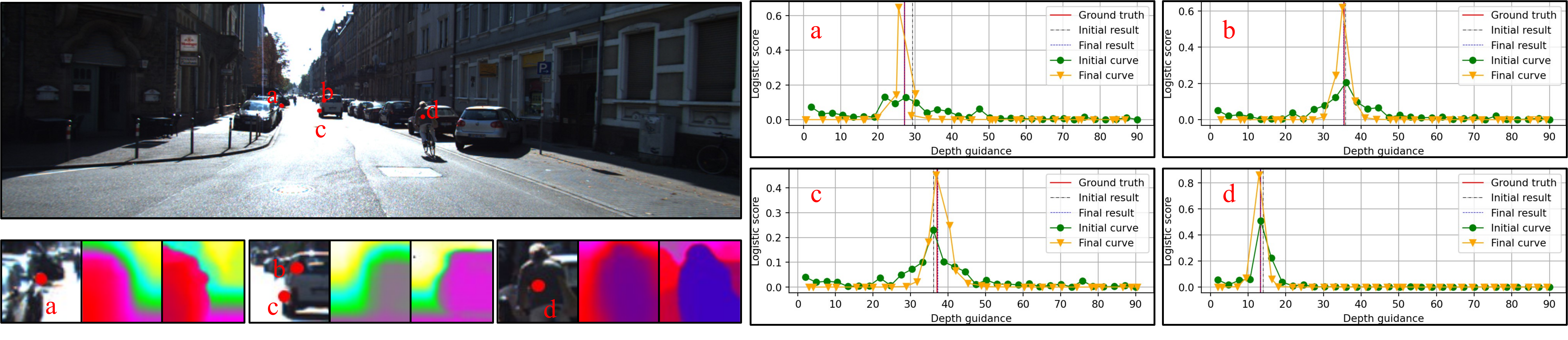}
\caption{The effect of the hierarchical translation module. We compare initial depth result ($\frac{1}{8}$ scale) and final depth result (original scale) at three different regions, and show logistic scores and depth guidance at four different points. After incorporating depth offsets, the logistic score curve is moved toward the ground truth, effectively addressing the issue of depth smearing.}
\label{fig_cur}
\end{figure*}
{\setlength{\parindent}{0cm} \textbf{Prune Translation Block}.} The block is designed to prune noise in the sparse depth map. As illustrated in Fig. \ref{fig_filter}, noise is typically present in the regions where the foreground and background overlap due to occlusion. Both the bicycle and signpost exhibit significant noise in depth measurement, which not only interfere with the geometric structure of the environment but also leads to incorrect depth cue.

Given that the estimated depth map exhibits a more reasonable and comprehensive object structure compared to the sparse depth map, we can utilize it as a reference to filter out the noise. In the filter, we first calculate the residual $R$ between the two depth maps, 
\begin{equation}
	R \left(x, y\right) = 
	\begin{cases}
		S_4\left(x, y\right) - D_4\left(x, y\right) & S_4\left(x, y\right)>0 \\
		0 & otherwise
	\end{cases},
\end{equation}
where $S_4$ is the original sparse depth map and $D_4$ is the estimated depth map at the original scale. If there is a zero value at $(x, y)$ in the sparse depth map, we set the corresponding residual to zero. Then, we introduce a threshold $\tau$ to distinguish between inliers and outliers. The mask $M_{in}$ denotes the inliers and is defined as $M_{in} \left(x, y\right) = \mathbbm{1} \left[|R \left(x, y\right)| < \tau \right]$, where $\mathbbm{1}$ is an indicator function. If the input is True, the output is 1. Otherwise, it is 0. Finally, the filter outputs the corrected sparse depth map $S_5 = S_4 \cdot M_{in}$, which contains all inliers and excludes the noise from outliers. As shown in Fig. \ref{fig_filter}, the structures of bicycle and signpost become distinct in the corrected sparse depth map. 

In the prune translation block, we use the corrected sparse depth map to adjust depth representation at the original scale to refine final depth prediction $D_5$. Here, we directly use the logistic score $L_4$ to reduce computation. 

With the hierarchical translation module, CluDe progressively adjust the depth representation to refine depth result. Fig. \ref{fig_cur} illustrates the initial and final depth prediction results. By incorporating both global and local depth offsets, we can continuously move the logistic score curve along the depth axis toward the ground truth by adjusting depth guidance, resulting in more clear edges around objects.

\subsection{Loss Function}
To train the entire network, we employ a loss function comprising three losses. Additionally, before calculating the loss, all predictions at lower scales are upsampled to the original scale. The first loss of the loss function is used to constrain the logistic score, we calculate cross-entropy (CE) with predicted logistic score and ground truth score followed \cite{chen2019over,lee2021depth},
\begin{equation}
	{\mathcal L}_1 = - \sum_{i \in \{1,2,3,4\}} \frac{1}{\left| \Omega \right|} \sum \hat{l} \log{l_i},
\end{equation}
where $\Omega$ represents the set of valid pixels, $l_i$ is the $i$th predicted logistic score, and $\hat{l}$ is the ground truth score generated by initial depth guidance via an approximate Laplace distribution \cite{chen2019over}. The other two losses of the loss function consist of the mean absolute error (MAE) and the mean squared error (MSE), respectively.
\begin{equation}
	\begin{split}
		&\mathcal L_2 = \sum_{i \in \{1,2,3,4,5\}} \frac{1}{\left| \Omega \right|} \sum \Vert \hat{D} - D_i \Vert_1  , \\
		&\mathcal L_3 = \sum_{i \in \{1,2,3,4,5\}} \frac{1}{\left| \Omega \right|} \sum \Vert \hat{D} - D_i \Vert_2^2,
	\end{split}
\end{equation}
where $D_i$ is the $i$th depth prediction, and $\hat{D}$ is the ground truth depth map.

The total loss function is the sum of the three aforementioned losses, denoted as $\mathcal L = \mathcal L_1 + \mathcal L_2 + \mathcal L_3$.
Thanks to the feature aggregation in the forward processing, and backward propagation based on the loss function, our method effectively generates a complete depth result guided by the RGB image.

\section{EXPERIMENTS}
We perform extensive experiments to validate the effectiveness of our method using both outdoor and indoor datasets captured in real-world scenes. We first introduce the experimental setup, which includes datasets, evaluation metrics used in our experiments and implementation details of our method in subsection \ref{es}. Then, we conduct comprehensive ablation studies to investigate the impact of each network module in subsection \ref{ab}. Next, we compare our method with typical classification-based methods and other state-of-the-art methods in subsection \ref{sota}. Finally, we present a comprehensive discussion of our method in subsection \ref{dis}.

\subsection{Experimental Setup} 
\label{es}
{\setlength{\parindent}{0cm}\textbf{KITTI Depth Completion}.} This dataset \cite{geiger2012we} is collected from an outdoor environment using a driving vehicle. It comprises 1216 $\times$ 352 sparse depth maps and their corresponding RGB images. The dataset is composed of 85898 training samples, 1000 selected validation samples, and 1000 test samples without ground truth depth maps. These test samples need to be submitted to the official KITTI website for evaluation. The sparse depth map is the raw output from the Velodyne LiDAR sensor, and the ground truth depth map is obtained by accumulating 11 consecutive scans, each with about 30\% annotated image points. 

{\setlength{\parindent}{0cm}\textbf{VOID}.} This dataset \cite{wong2020unsupervised} is captured in both indoor and outdoor environments using an off-the-shelf VIO system driven by an Intel RealSense D435i camera. It contains synchronized 640 $\times$ 480 sparse depth maps and RGB images of 56 indoor and outdoor scenes, including classrooms, offices, stairwells, laboratories, and gardens. The dataset consists of 48248 training samples and 800 selected validation samples. The sparse depth maps in the dataset are provided at different densities. Specifically, there are sparse depth maps with 1500, 500, and 150 points, corresponding to densities of 0.5\%, 0.15\%, and 0.05\% respectively. For training our network, we utilize the sparse depth maps containing 1500 points along with their corresponding RGB images as training samples.

\begin{table}[h]
\centering
\caption{Definition and unit of evaluation metrics.}
\begin{tabular}{ccc}
\hline
Metric     & Define &  Unit \\
\hline 
MAE  & $\frac{1}{\left| \Omega \right|} \sum \left| \hat{D} - D \right|$ & mm  \\
RMSE & $\left( \frac{1}{\left| \Omega \right|} \sum \left| \hat{D} - D \right|^2 \right)^\frac{1}{2}$ & mm  \\
iMAE & $\frac{1}{\left| \Omega \right|} \sum \left| \frac{1}{\hat{D}} - \frac{1}{D} \right|$ & 1/km\\
iRMSE& $\left( \frac{1}{\left| \Omega \right|} \sum \left| \frac{1}{\hat{D}} - \frac{1}{D} \right|^2 \right)^\frac{1}{2}$ & 1/km\\
Latency & The average runtime & Second \\
FLOPs & Floating point of operations & Gillion \\
Param. & The parameters of the network & Million \\
\hline
\end{tabular}
\label{tab_m}
\end{table}

{\setlength{\parindent}{0cm}\textbf{Metrics}.} Similar to previous works \cite{liu2021fcfr,peng2022pixelwise,wong2020unsupervised}, we use four standard metrics to evaluate the accuracy, which are mean absolute error (MAE), root mean squared error (RMSE), mean absolute error of the inverse depth (iMAE) and root mean squared error of the inverse depth (iRMSE). In addition, we report the latency, FLOPs and parameters of the network. TABLE \ref{tab_m} summarizes the evaluation metrics, including their respective definitions and units.

{\setlength{\parindent}{0cm}\textbf{Implementation Details}.} We use PyTorch to implement our method and AdamW optimizer with a weight decay of 0.05 to train the whole network on a single Nvidia A40 GPU. We adopt a two-stage training strategy to train the network. In the first stage, we train the network except for the prune translation block for 60 epochs, and set the learning rate to \{5e-4, 4e-4, 3e-4, 2e-4, 1e-4, 5e-5, 1e-5, 1e-6\} at epoch \{1, 20, 25, 30, 35, 45, 50, 55\}. In the second stage, we only train the prune translation block for 16 epochs, and cyclically set the learning rate to \{5e-4, 4e-4, 3e-4, 2e-4, 1e-4\} in the first ten epochs and \{5e-5, 1e-5\} in the remainder. 

For the KITTI dataset, we first crop the images from the bottom to a size of 1216 $\times$ 320 to eliminate the region that lacks depth information. Subsequently, we randomly crop the images to a size of 738 $\times$ 256 and set the batch size to 8. Additionally, we use horizontal flipping and color jitter to augment the data for training. Regarding other hyperparameters, we set the maximum depth to 90m, and $K$ is set to 32. The depth guidance is initialized by uniformly sampling $K$ values. The threshold $\tau$ in the filter of the prune translation block is empirically set to 0.5m. For the VOID dataset, we randomly crop images to 512 $\times$ 384, and utilize a batch size of 16 to train the network, using the same data augmentation as KITTI. We cap the depth values between 0.2m and 5.0m following the evaluation protocol of the work \cite{liu2022monitored}, and set $K$ to 16 like \cite{kam2022costdcnet}. The depth guidance is also initialized by uniformly sampling 16 values. The threshold $\tau$ using in the filter of the prune translation block is empirically set to 0.25m.

Following these settings, we train the default model CluDe with all losses in the loss function assigned the same weight. Furthermore, we train CluDe$^{\dagger}$ model, adjusting the weight of the loss $\mathcal{L}_3$ to 0.2 for improved results on the KITTI.

\subsection{Ablation Study}
\label{ab}
We conduct ablation experiments to illustrate the effectiveness of the proposed CluDe. We use the KITTI dataset to evaluate the effect of different settings, and VOID to validate the robustness to different input densities.

\begin{table}[h]
\centering
\scriptsize
\caption{Comparison of different frameworks. CLS denotes classification-based method. CLT denotes our clustering-based method. Num. denotes the number of categories. The lower the metrics, the better.}
\begin{tabular}{cccccccc}
\hline
Method  & Offset & Num.   & MAE & RMSE & iMAE & iRMSE & Param. \\
\hline
CLS & -          & 32     & 382.97& 1105.53  & 1.91& 4.10 & 19 \\
CLT & -          & 32     & 391.62& 1143.24  & 1.74& 2.88 & 21 \\ 
CLS & \checkmark & 32     & 214.99& 817.11   & 0.93& 2.42 & 20 \\
CLT & \checkmark & 32     & 213.73& 812.21   & 0.93& 2.37 & 22 \\
CLT & \checkmark & 16     & 231.99& 834.48   & 1.06& 2.64 & 22 \\
\hline
\end{tabular}
\label{tab_cc}
\end{table}

{\setlength{\parindent}{0cm}\textbf{Clustering vs. Classification}.} To compare our clustering-based method with the classification-based method, we remove the learnable depth centers in the clustering transformer to construct a typical classification-based framework. In this framework, the transformer no longer performs clustering but only propagates depth information between long-range tokens. Firstly, we compare two methods that do not incorporate the hierarchical translation module for estimating depth offset. According to the results presented in TABLE \ref{tab_cc}, our clustering-based method exhibits inferior performance compared to the classification-based method. This can be primarily attributed to the fact that the initial depth centers, represented by a set of learnable feature vectors, require more fine-grained guidance to capture the details of the depth distribution. After incorporating the hierarchical translation module to estimate depth offsets, our clustering-based method gradually refines the depth guidance to provide more fine-grained hints for learning the depth centers. As a result, the performance of our method surpasses that of the conventional classification-based method. Moreover, depth offsets improve the performance of both methods, further demonstrating the significance of pixel-wise and continuous depth representation.

We also validate the impact of the number of depth centers on limited GPU memory in TABLE \ref{tab_cc}. It is observed that increasing the number of depth centers enables a more accurate representation of the depth range, thereby leading to better performance.

\begin{table}[h]
\centering
\scriptsize
\caption{The effect of hierarchical translation module. Up. denote upsampling operation. The lower the metrics, the better.}
\begin{tabular}{cccccccc}
\hline
Global & Local & Up.          & MAE & RMSE & iMAE & iRMSE & Param.\\
\hline
-          & -          & -          & 391.62 & 1143.24& 1.74& 2.88 & 21   \\
\checkmark & -          & \checkmark & 282.43 & 862.91 & 1.63& 3.31 & 22  \\
-          & \checkmark & \checkmark & 227.31 & 847.12 & 0.97& 2.53 & 22  \\
\checkmark & \checkmark & \checkmark & 215.87 & 814.56 & 0.93& 2.39 & 22  \\
\checkmark & \checkmark & -          & 233.78 & 848.68 & 1.05& 2.69 & 22  \\
\hline
\end{tabular}
\label{tab_oo}
\end{table}

{\setlength{\parindent}{0cm}\textbf{Continuous depth representation}.} 
In our method, the depth centers generated by clustering remain discrete. To enhance the continuity of this depth representation, we introduce the hierarchical translation module. This module can estimate depth offsets to adjust depth guidance which provides fine-grained hints for the depth centers. As shown in Fig. \ref{fig_cur}, the depth offsets induce a shift of the logistic score curve along the depth axis to promote continuity. We report the performance of different settings in TABLE \ref{tab_oo}. Global depth offset allows moving the whole logistic score curve without any depth range constraints to improve global continuity, and local depth offsets fine-tune each one in a narrow depth band to search for a local continuous representation. Simultaneously estimating both depth offsets can significantly enhance continuity of depth representation and improve overall performance. In Fig. \ref{fig_cur}, the initial logistic score curve leads to ambiguous depth result. Offsets adjust the depth guidance towards ground truth to generate a continuous curve and solve the problem of depth smearing.

In addition, we employ upsampling on the depth feature $\tilde{F}$ to gradually restore the original scale depth result. As a comparison, we directly upsample the output of the module to obtain the depth result at the original scale. The results presented in TABLE \ref{tab_oo} demonstrate that gradually upsampling on the depth feature yields superior performance, since this operation can restore more accurate depth information, leading to improved results.

\begin{table}[h]
\centering
\footnotesize
\caption{The effect of depth cue. PTB denotes the prune translation block}
\begin{tabular}{cccccc}
\hline
Depth cue       & PTB   & MAE $\downarrow$  & RMSE $\downarrow$  & iMAE $\downarrow$ & iRMSE $\downarrow$ \\
\hline
-          & -          & 252.75 & 855.53 & 1.15  & 2.78  \\
\checkmark & -          & 213.73 & 812.21 & 0.93  & 2.37  \\
\checkmark & with filter& 206.04 & 803.60 & 0.89  & 2.30  \\
\checkmark & w/o filter & 207.08 & 804.62 & 0.90  & 2.30  \\
\hline
\end{tabular}
\label{tab_dc}
\end{table}

{\setlength{\parindent}{0cm}\textbf{The effect of depth cue}.} Another crucial aspect of the hierarchical translation module is the depth cue used to estimate depth offsets. These depth cues capture the residual between sparse depth map and depth result. Although these cues are sparse, they provide essential guidance for depth offset estimation by indicating the direction of prior adjustment. The results presented in TABLE \ref{tab_dc} demonstrate the significant role of these depth cues. This exploration demonstrates an alternative use of the sparse depth map, beyond its traditional role as an input.

\begin{figure}[h]
\centering
\includegraphics[width=0.9\linewidth]{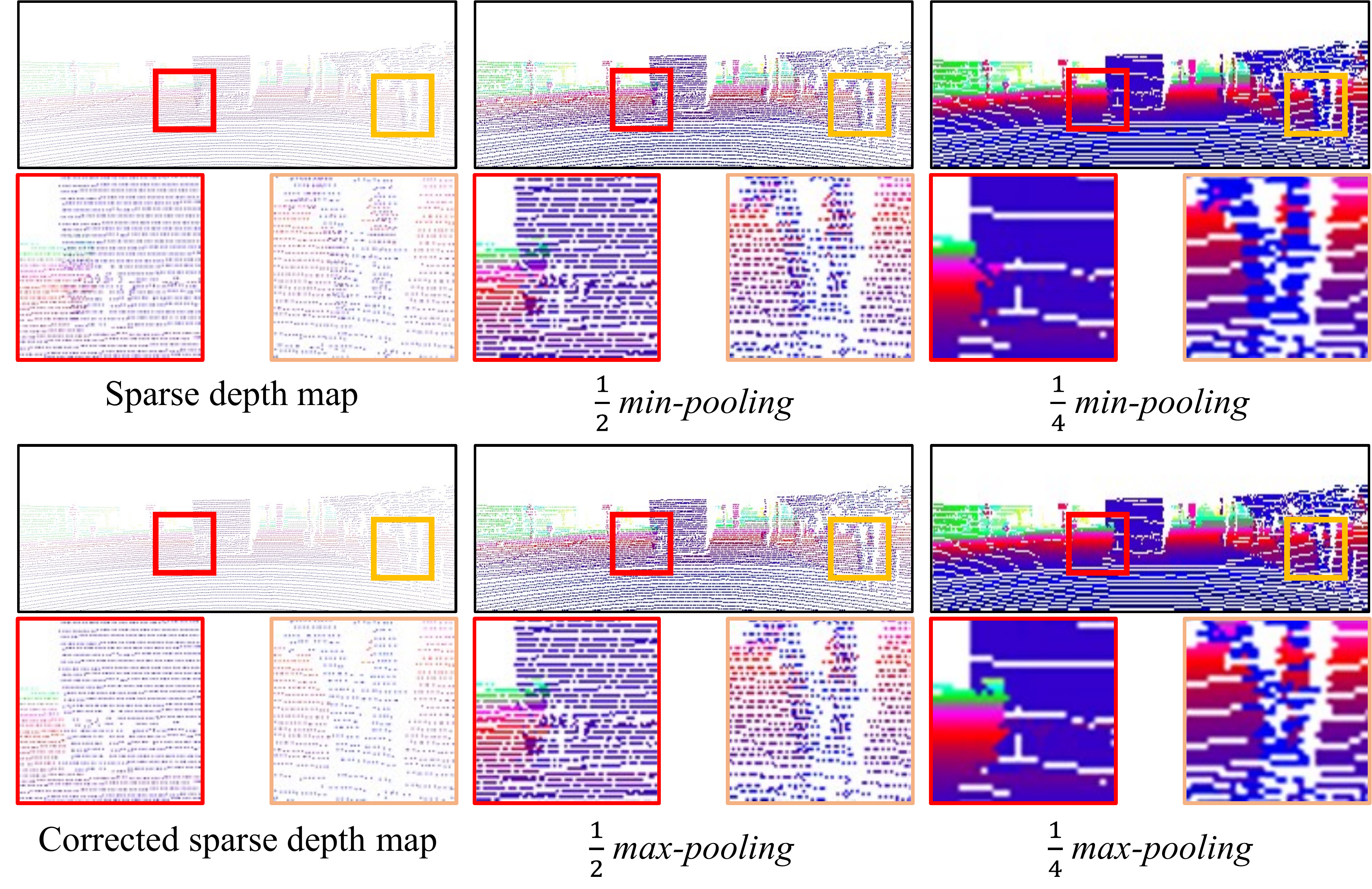}
\caption{Visualization of multi-scale sparse depth maps. We zoom in the highlighted region for better visibility.}
\label{fig_cue}
\end{figure}

Furthermore, we observe that noise in the sparse depth map can introduce erroneous depth cue. To address this issue, we employ pooling operations to filter out noise in the sparse depth map, except for the original scale. In Fig. \ref{fig_cue}, we showcase the results at $\frac{1}{2}$ and $\frac{1}{4}$ scale, demonstrating that \textit{max-pooling} and \textit{min-pooling} operations can effectively mitigates the noise. For the original scale, we introduce a simple filter integrated into our prune translation block, as there is only a small amount noise present at the boundary. The corrected sparse depth map in Fig. \ref{fig_cue} contains less noise than the sparse depth map, resulting in improved performance.

{\setlength{\parindent}{0cm}\textbf{The efficiency of each module}.}
TABLE \ref{tab_iter} reports accuracy, latency and FLOPs of different settings. If the hierarchical translation module is removed, our method comprises only an encoder and a clustering transformer. After introducing the hierarchical translation module, accuracy significantly improves due to the module providing fine-grained depth guidance for depth representation. Furthermore, we observe that accuracy improves with an increase in iterative rounds. The full hierarchical translation module results in over 100\% relative improvement in MAE (from 480 to 206) with 0.06s additional runtime cost per image. These results demonstrate that our modules are indispensable in learning pixel-wise and continuous depth representation.

\begin{table}[h]
\centering
\scriptsize
\caption{Comparison of efficiency in different settings. EC: Encoder and Clustering transformer. ECH: EC and Hierarchical translation module. The lower the metrics, the better.}
\begin{tabular}{cc|ccccccc}
\hline
Setting& Round & MAE & RMSE & iMAE & iRMSE  & Latency & FLOPs \\
\hline
EC & - & 480.04 & 1373.48& 2.48 & 5.17 & 0.08 & 123 \\
ECH &1   & 395.45 & 1147.63& 2.19 & 4.66 & 0.08 & 127 \\
ECH &2   & 299.80 & 1028.62& 1.34 & 3.59 & 0.09 & 148 \\  
ECH &3   & 215.89 & 814.52 & 0.94 & 2.50 & 0.11 & 223 \\ 
ECH &4   & 206.03 & 803.60 & 0.89 & 2.30 & 0.14 & 254 \\
\hline
\end{tabular}
\label{tab_iter}
\end{table}

{\setlength{\parindent}{0cm}\textbf{The effect of loss function}.}
The loss function used in our method comprises the CE loss (${\mathcal L}_1$), MAE loss (${\mathcal L}_2$), and MSE loss (${\mathcal L}_3$), and final loss is formulated as ${\mathcal L}_1 + {\mathcal L}_2 + {\mathcal L}_3$. To quantify their effects, we provide an ablation study for loss items. As shown in TABLE \ref{tab_loss}, we can find that a single CE loss is insufficient to learn stable depth centers, leading to the worst performance. When the CE loss is combined with other losses, the performance will be significantly improved. The MAE loss and MSE loss can largely reduce the depth errors based on corresponding metrics. This may result in some loss function combinations outperforming our final solution on some specific metrics. Based on the consideration of improving the overall performance, our CluDe still adopts all three loss items. Furthermore, our CluDe$^{\dagger}$ adjusts the weight of MSE loss to improve the MAE metric.

\begin{table}[h]
\centering
\caption{Comparison of different loss functions.}
\begin{tabular}{ccc|cccc}
\hline
\multicolumn{3}{c|}{Loss} &  \multicolumn{4}{c}{Metric} \\
CE & MAE & MSE & MAE $\downarrow$& RMSE $\downarrow$& iMAE $\downarrow$& iRMSE $\downarrow$ \\
\hline
\checkmark& -    &  -           & 808.61 & 1597.05& 5.09 & 8.54 \\
-&\checkmark& -                 & 216.35 & 926.68 & 0.93 & 2.67 \\
-& -    &\checkmark             & 275.66 & 925.58 & 1.35 & 3.54  \\
\checkmark&\checkmark&-         & 208.75 & 866.53 & 0.91 & 2.36 \\
\checkmark&-&\checkmark         & 229.43 & 828.86 & 1.01 & 2.50 \\
-&\checkmark&\checkmark         & 228.22 & 837.61 & 1.01 & 2.61 \\
\checkmark&\checkmark&\checkmark& 215.89 & 814.52 & 0.94 & 2.50 \\
\hline   
\end{tabular}
\label{tab_loss}
\end{table}

\begin{figure}[h]
\centering
\includegraphics[width=0.9\linewidth]{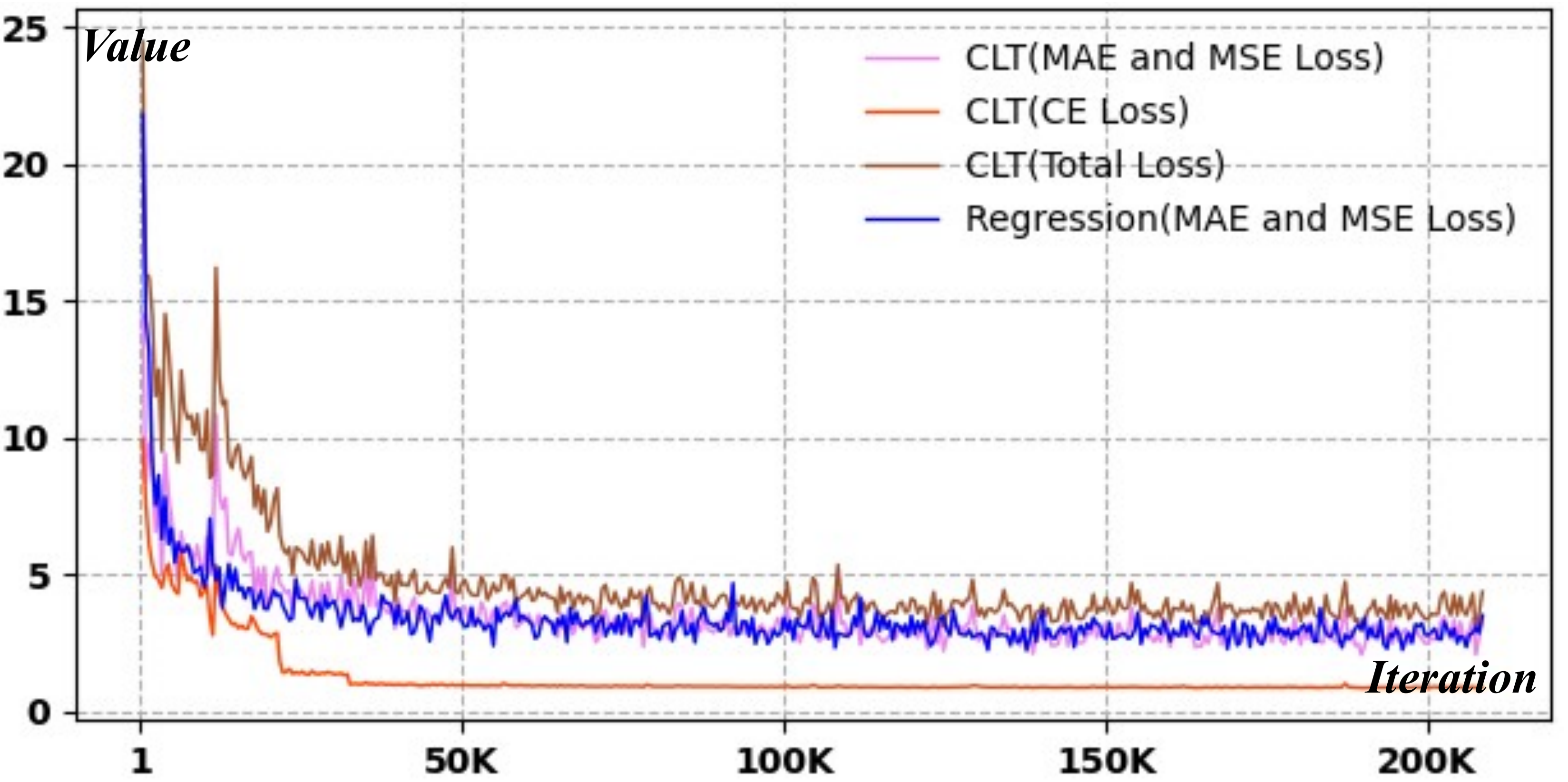}
\caption{The losses of clustering-based (CLT) and regression-based (Regression) frameworks.}
\label{fig_conv}
\end{figure}

{\setlength{\parindent}{0cm}\textbf{The convergence speed of our framework}.}
We conduct this experiment to compare the convergence speed of our clustering-based framework with the regression-based framework. We first replace the classifier in CluDe with a regressor to build the corresponding regression-based variant. The loss function used to train the variant comprises the CE loss and MAE loss, while CluDe incorporates the MAE loss, MSE loss, and CE loss. 	
In Fig. \ref{fig_conv}, while the shapes of MAE and MSE losses across different models are almost consistent, the CE loss converges more rapidly, facilitating convergence for our CluDe. As shown in TABLE \ref{tab_con}, we observe that the regression-based variant exhibits instability in the initial phase, as indicated by the disrupted iMAE and iRMSE. At the 140K iteration, both models achieve their best performance, and our method is better due to its quicker convergence speed. In subsequent iterations, they exhibit indications of overfitting. These results demonstrate that the clustering-based method achieves a superior convergence speed compared to the regression-based method with the same architecture.


\begin{table}[h]
\centering
\caption{The performance of clustering-based (CLT) and regression-based (Regression) frameworks at the same iteration.}
\begin{tabular}{cc|cccc}
\hline
Method & Iteration & MAE $\downarrow$& RMSE $\downarrow$& iMAE $\downarrow$& iRMSE $\downarrow$ \\
\hline
CLT&  10K   & 260.12 & 977.98 & 1.16 & 3.73      \\
CLT&  70K   & 226.44 & 848.02 & 0.98 & 2.47       \\
CLT&  140K  & 216.87 & 812.23 & 0.94 & 2.38       \\
CLT&  210K  & 220.67 & 815.36 & 0.99 & 2.45       \\
\hline
Regression&  10K   & 258.23 & 949.86 & - & - \\
Regression&  70K   & 226.62 & 840.58 & 0.99 & 2.54  \\
Regression&  140K  & 220.72 & 824.29 & 0.97 & 2.47 \\
Regression&  210K  & 248.03 & 827.44 & 1.25 & 2.81 \\
\hline   
\end{tabular}
\label{tab_con}
\end{table}

\begin{figure}[h]
\centering
\includegraphics[width=0.9\linewidth]{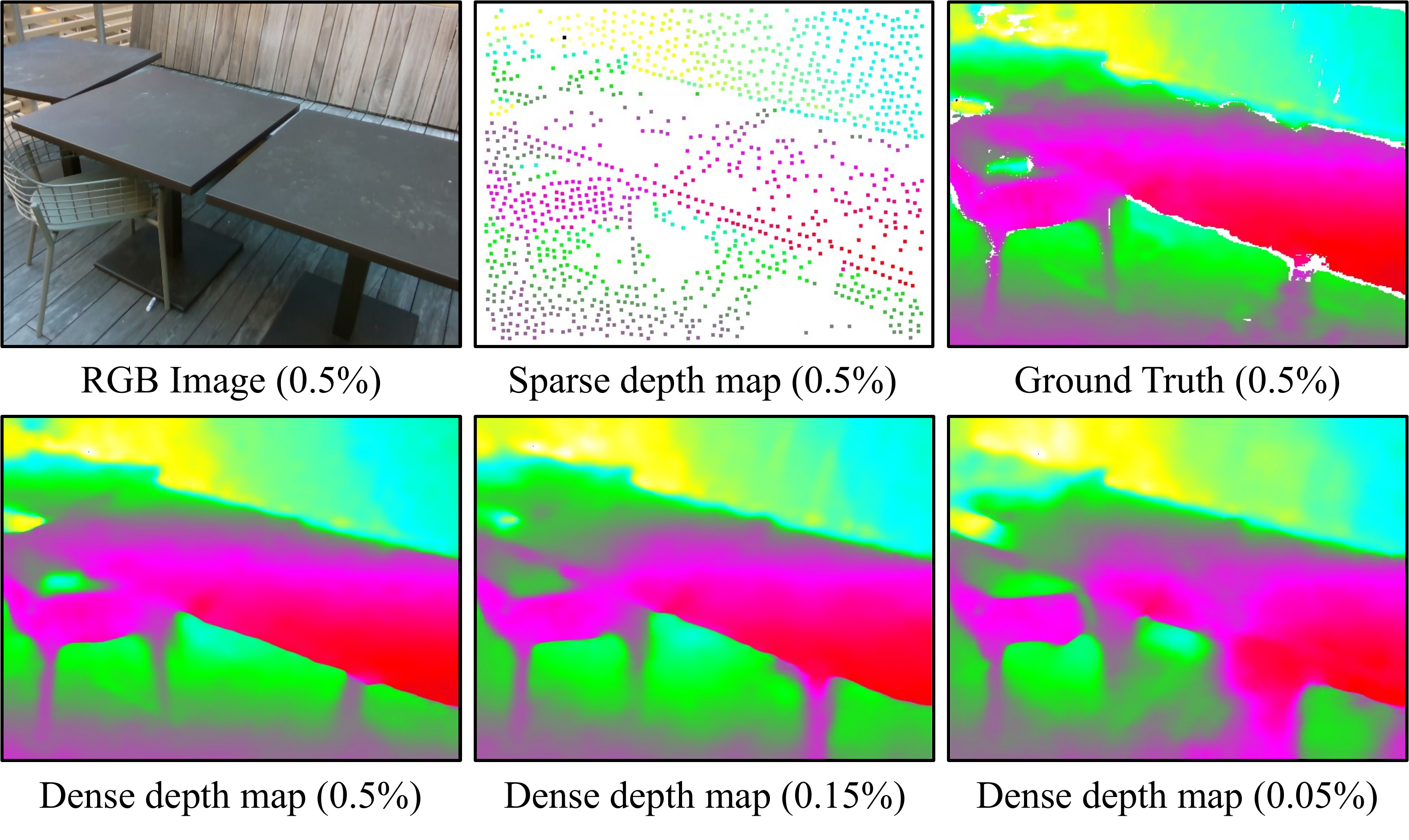}
\caption{Dense depth maps of different input densities. As the density of sparse depth map decreases, the quality of the depth results tends to deteriorate.}
\label{fig_void}
\end{figure}

{\setlength{\parindent}{0cm}\textbf{Robustness to density of sparse depth map}.} To evaluate the robustness of CluDe to different input densities, we test CluDe using the sparse depth map with 500 and 150 valid depth values which take only 0.15\% and 0.05\% of the dense depth map. As reported in TABLE \ref{tab_vo}, despite a decrease in performance as the density decreases, CluDe is still capable of producing reasonable depth results. When changing the sparse input to 150 values, CluDe still attains a result of about 168mm evaluated by RMSE, which demonstrates CluDe is robust well to different input densities. Due to the inconsistency of data with different densities in the VOID dataset, we only show one group depth result at the approximate scene in Fig. \ref{fig_void}. 

\begin{table}[h]
\centering
\caption{Robustness of density on VOID test dataset.}
\begin{tabular}{ccccc}
\hline
Density  & MAE $\downarrow$   & RMSE  $\downarrow$ & iMAE $\downarrow$ & iRMSE $\downarrow$\\
\hline
150      & 85.48  & 168.90 & 43.03 & 86.92 \\
500      & 53.25  & 120.39 & 25.50 & 60.00 \\
1500     & 25.89  & 72.35  & 12.91 & 34.15 \\
\hline
\end{tabular}
\label{tab_vo}
\end{table}

\begin{table}[h]
\centering
\scriptsize
\caption{Comparison of different methods at boundary and non-boundary on KITTI validation set. Imp. is the improvement on MAE.} 
\begin{tabular}{l|ccccc}
\hline
Method      & MAE $\downarrow$ & RMSE  $\downarrow$ & iMAE $\downarrow$ & iRMSE $\downarrow$ & Imp. $\uparrow$\\
\hline 
\multicolumn{6}{c}{boundary} \\
\hline
NLSPN  \cite{park2020non}     & 377.91  & 1323.77 & 0.95 & 2.88 & 4.23\% \\
PENet \cite{hu2020PENet}      & 382.36  & 1249.34 & 1.01 & 2.88 & 5.63\% \\
ACMNet\cite{zhao2021adaptive} & 392.70  & 1262.63 & 1.04 & 2.89 & 7.82\% \\
CLS                           & 379.35  & 1272.48 & 0.97 & 2.82 & 4.58\% \\
\hline 
CluDe (ours)                         & \textbf{361.98} & \textbf{1248.79} & \textbf{0.92} & \textbf{2.79} & -\\
\hline \hline
\multicolumn{6}{c}{non-boundary} \\
\hline 
NLSPN  \cite{park2020non}     & 185.87 & 711.09 & 0.86 & 2.02 & 2.34\% \\
PENet \cite{hu2020PENet}      & 191.10 & 676.65 & 0.92 & 2.13 & 5.01\% \\
ACMNet\cite{zhao2021adaptive} & 194.95 & 689.62 & 0.92 & 2.10 & 6.89\% \\
CLS                           & 187.70 & 691.84 & 0.88 & 2.05 & 3.29\% \\
\hline 
CluDe (ours)                        & \textbf{181.52} & \textbf{682.53} & \textbf{0.86} & \textbf{1.99} & -\\
\hline
\end{tabular}
\label{tab_b}
\end{table}

\begin{table*}[t]
	\centering
	\caption{Comparing MAE of different methods across various depth intervals on the KITTI validation set. CLS is the classification-based depth completion method. The lower the metrics, the better.}
	\begin{tabular}{c|ccccccccc}
		\hline
		Range                                          & (0, 5)  & (5, 10)  & (10, 15) & (15, 20) & (20, 25) & (25, 30) & (30, 35) & (35, 40) & (40, 45) \\
		\hline
		NLSPN  \cite{park2020non}     & \textbf{104.12}  & \textbf{70.53}    & \underline{121.62}   & \underline{192.12}   & \underline{299.50}   & \underline{430.15}   & 567.08   & \underline{737.53}   & \underline{930.56}   \\
		PENet \cite{hu2020PENet}      & 158.15  & 73.92    & 128.33   & 205.80   & 317.28   & 450.89   & 601.05   & 780.65   & 970.22   \\
		ACMNet\cite{zhao2021adaptive} & 151.84  & 74.40    & 125.31   & 198.25   & 310.67   & 442.89   & 590.64   & 756.43   & 944.07   \\
		CLS                                            & 136.69  & 71.03    & 122.02   & 198.36   & 300.73   & 436.16   & \underline{563.76}   & 739.24   & 948.48   \\
		\hline
		CluDe (ours)                                   & \underline{104.59}  & \underline{70.94}    & \textbf{119.44}   & \textbf{188.20}   & \textbf{294.76}   & \textbf{414.12}   & \textbf{551.77}   & \textbf{702.94}   & \textbf{895.34}   \\
		\hline\hline
		Range                                          & (45, 50) & (50, 55) & (55, 60) & (60, 65) & (65, 70) & (70, 75) & (75, 80) & (80, 85) & (85, 90) \\
		\hline
		NLSPN  \cite{park2020non}     & \underline{1094.44} & 1282.18  & 1461.98  & 1592.48  & 1843.16  & 2205.92  & 2658.23  & 6386.87  & 11966.13 \\
		PENet \cite{hu2020PENet}      & 1147.97 & 1307.72  & 1479.99  & 1617.96  & 1837.15  & 2059.33  & \textbf{2197.45}  & 3854.77  & \underline{5988.07}  \\
		ACMNet\cite{zhao2021adaptive} & 1113.84 & \underline{1258.27}  & \underline{1411.15}  & \underline{1536.26}  & \textbf{1742.76}  & \textbf{1977.53}  & \underline{2235.17}  & \underline{4324.39}  & 8679.69  \\
		CLS                                            & 1111.88 & 1304.92  & 1485.13  & 1635.27  & 1910.86  & 2182.47  & 2409.90  & \textbf{4196.45}  & \textbf{5431.17}  \\
		\hline
		CluDe (ours)                                   & \textbf{1052.04} & \textbf{1240.99}  & \textbf{1397.80}  & \textbf{1526.64}  & \underline{1755.12}  & \underline{1987.08}  & 2316.30  & 4387.26  & 7373.00  \\
		\hline
	\end{tabular}
	\label{tab_s}
\end{table*}

\subsection{Comparison with State-of-the-Art Methods}
\label{sota}
\begin{figure}[h]
	\centering
	\includegraphics[width=0.9\linewidth]{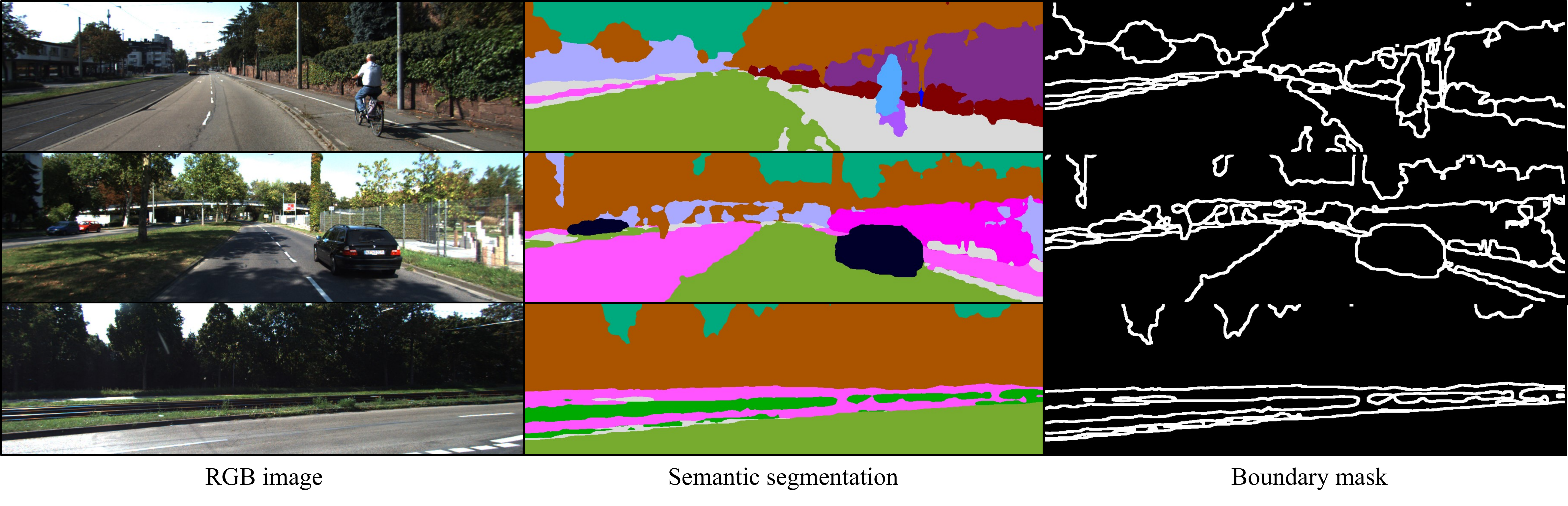}
	\caption{Visualization of semantic segmentation and boundary mask.}
	\label{fig_boundary}
\end{figure}

{\setlength{\parindent}{0cm}\textbf{Improvement at boundary and non-boundary}.} To demonstrate the effectiveness of the pixel-wise and continuous depth representation constructed by CluDe in capturing the real depth distribution and addressing depth smearing at object boundaries, we evaluate the performance of different methods on both boundary and non-boundary regions within the KITTI validation set. To generate the boundary, we first calculate the semantic segmentation result for an RGB image \cite{Xu_2023_CVPR}. Next, we identify pixels with different semantics in the neighboring region as the boundary region and expand this region with a $5 \times 5$ window, while the remaining pixels are considered the non-boundary region. Fig. \ref{fig_boundary} showcases several examples of boundary masks, which, although somewhat rough, still provide a reference for the boundaries. 

We compare CluDe with NLSPN \cite{park2020non}, PENet \cite{hu2020PENet}, ACMNet \cite{zhao2021adaptive} and a classification-based depth completion method, and report performance in TABLE \ref{tab_b}. CluDe exhibits superior performance in both boundary and non-boundary regions, with particularly notable improvements observed in the boundary regions. Conventional classification-based methods rely on pixel-shared and discrete depth categories to represent the depth distribution, which fails to capture continuous depth values. In contrast, CluDe learns depth centers and offsets from the dataset, enabling pixel-wise continuous depth representation, which significantly enhances performance, particularly in boundary regions.

\begin{figure}[h]
\textsc{\centering
\includegraphics[width=0.9\linewidth]{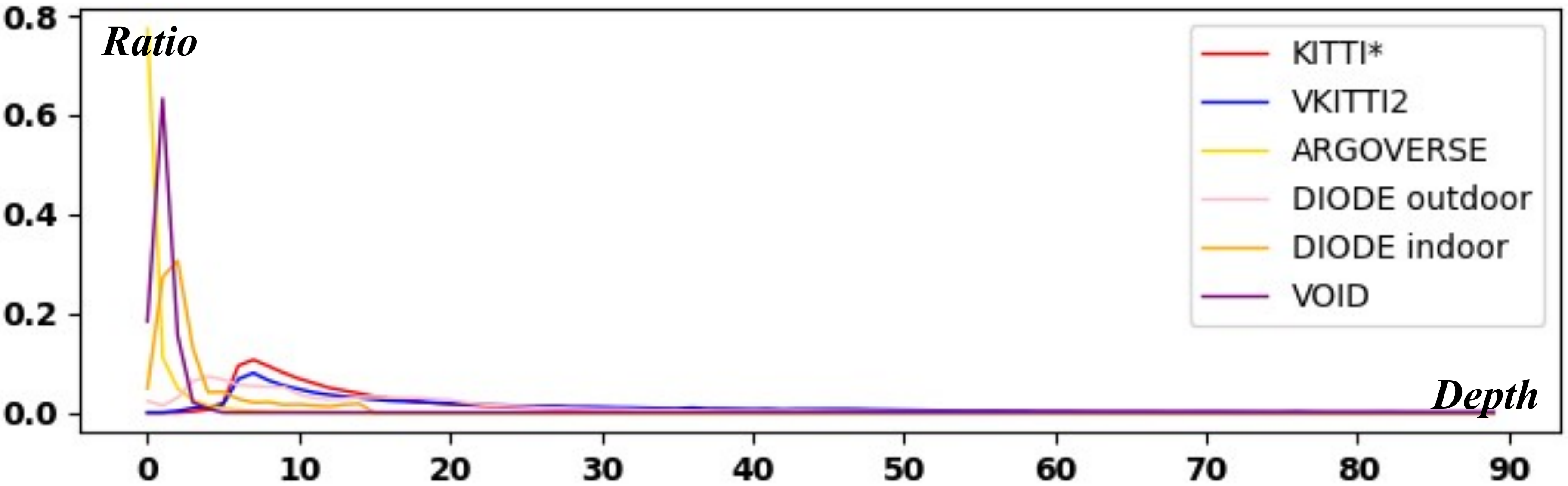}
\caption{The depth distribution of different datasets. $*$ denotes the training dataset.}
\label{fig_dis}}
\end{figure}

{\setlength{\parindent}{0cm}\textbf{Performance at different depth range}.} To comprehensively evaluate our proposed CluDe and analyze the learned depth centers, we report the performance of different methods across various depth ranges. TABLE \ref{tab_s} presents the MAE for different depth intervals within 90m. As the distance increases, the MAE gradually increases as well. Our method excels in the medium depth range. If the depth exceeds 75m, all methods exhibit poor performance. Fig. \ref{fig_dis} illustrates that the depth distribution of most datasets follows a long-tail pattern. For the KITTI dataset, only 0.87\% of the valid points have depths exceeding 65m, and a mere 0.04\% exceed 80m. Compared to the conventional classification-based method, our CluDe achieves superior performance for distances up to 80m, since it cannot effectively learn depth representations to cover the entire depth range. However, when the depth extends beyond 65m, both the classification-based method and our method show inferior performance compared to several regression-based methods at specific depth intervals. This observation suggests that accurately predicting depth result for categories closer to the tail end of the depth distribution remains a challenge.

{\setlength{\parindent}{0cm}\textbf{Comparison with classification-based methods}.} As presented in TABLE \ref{tab_k}, we report the performance of our method and previous outstanding classification-based methods. Essentially, our CluDe is a classification-based method but exploits a clustering-based framework. Among these methods, CluDe stands out with its unique framework, which focuses on learning depth centers to represent the real depth distribution. 

\begin{table}[h]
\centering
\caption{Comparison of different classification-based methods on the outdoor and indoor scenes.}
\begin{tabular}{l|cccc}
\hline
Method      & MAE $\downarrow$  & RMSE $\downarrow$ & iMAE $\downarrow$  & iRMSE $\downarrow$  \\
\hline
\multicolumn{5}{c}{KITTI test set (outdoor)} \\
\hline
DC \cite{imran2019depth}          & 215.75 & 965.87 & 0.98 & 2.43 \\
PADNet \cite{peng2022pixelwise}   & 197.99 & 746.19 & \textbf{0.85} & \textbf{1.96} \\
\hline 
CluDe (ours)                              & 200.48 & \textbf{734.59} & 0.88 & 2.08  \\
CluDe$^{\dagger}$ (ours)                  & \textbf{197.91} & 742.26 & 0.86 & 2.02 \\
\hline  \hline
\multicolumn{5}{c}{VOID validation set (indoor and outdoor)} \\
\hline
CostDCNet \cite{kam2022costdcnet} & 25.84 & 76.28 & \textbf{12.19} & \textbf{32.13} \\
\hline 
CluDe (ours)                              & \textbf{25.74 } & \textbf{72.34 } & 12.76  & 33.92  \\
\hline
\end{tabular}
\label{tab_k}
\end{table}

On the KITTI dataset, CluDe surpasses DC \cite{imran2019depth} by a significant margin in all metrics and achieves comparable performance to PADNet \cite{peng2022pixelwise}. Despite DC pre-defining 80 depth categories, more than twice the number of CluDe, its inferior performance suggests that the number of depth categories is not the primary factor affecting performance. Instead, it is the allocation of these depth categories that plays a crucial role. Both CluDe and PADNet learn adaptive depth categories in the networks. CluDe learns pixel-wise continuous depth representation that effectively describes depth values across the entire depth range. In contrast, PADNet adjusts depth categories in the neighborhood of coarse result for refinement, but the output of PADNet is still based on the discrete depth representation. As shown in TABLE \ref{tab_k}, although CluDe performs 1.6\% better than PADNet in RMSE, it exhibits slightly lower performance in other metrics. When the weight of CE loss in the total loss function is decreased to 0.2, CluDe$^{\dagger}$ achieves state-of-the-art performance. However, it should be noted that both CluDe and CluDe$^{\dagger}$ perform worse than PADNet regarding iRMSE and iMAE. This is because our method focuses on learning depth centers that represent the entire depth range, rather than the local. As a result, the depth result close to a near depth center may be influenced by other distant depth centers, leading to inferior performance in these two metrics.

Unlike KITTI dataset, the data of VOID dataset only provides 1500 sparse depth values as the environment prior, and an abundance of zero values can detrimentally impact the depth result. Nonetheless, our method achieves the best RMSE and MAE compared to other methods. Similar to the observation on the KITTI dataset, CluDe exhibits slightly lower performance compared to CostDCNet \cite{kam2022costdcnet} in losses of iRMSE and iMAE metrics. This can be attributed to the fact that CluDe learns depth representation to effectively represent the entire real depth distribution and estimate the depth result near a certain depth center using all depth centers.

\begin{table}[h]
\centering
\caption{Comparison of different methods on the outdoor and indoor scenes.}
\begin{tabular}{c|cccc}
\hline
Method      & MAE $\downarrow$  & RMSE $\downarrow$  & iMAE $\downarrow$ & iRMSE $\downarrow$  \\
\hline 
\multicolumn{5}{c}{KITTI test set (outdoor)} \\
\hline
GAENet \cite{9811556}             & 231.29 & 773.90 & 1.08 & 2.29 \\
DeepLiDAR \cite{Qiu_2019_CVPR}    & 226.50 & 758.38 & 1.15 & 2.56 \\
GuideNet \cite{9286883}           & 218.83 & 736.24 & 0.99 & 2.25 \\
FCFR-Net \cite{liu2021fcfr}       & 217.15 & 735.81 & 0.98 & 2.20 \\
PENet \cite{hu2020PENet}          & 210.55 & 730.08 & 0.94 & 2.17 \\
GuideFormer \cite{rho2022guideformer} & 207.76 & 721.48 & 0.97 & 2.14 \\
ACMNet \cite{zhao2021adaptive}    & 206.09 & 744.91 & 0.90 & 2.08 \\
DGDF-A \cite{10173567}            & 205.01 & 708.30 & 0.91 & 2.04 \\
RigNet \cite{yan2021rignet}       & 203.25 & 712.66 & 0.90 & 2.08 \\
NLSPN \cite{park2020non}          & 199.59 & 741.68 & 0.84 & 1.99 \\
DySPN \cite{lin2022dynamic}       & 192.71 & 709.12 & 0.82 & 1.88 \\
\hline
CluDe (ours)                              & 200.48 & 734.59 & 0.88 & 2.08 \\
PADNet \cite{peng2022pixelwise}   & 197.99 & 746.19 & 0.85 & 1.96 \\
CluDe$^{\dagger}$ (ours)                  & 197.91 & 742.26 & 0.86 & 2.02 \\
\hline \hline
\multicolumn{5}{c}{VOID validation set (indoor and outdoor)} \\
\hline
PENet \cite{hu2020PENet}          & 34.60 & 82.01 & 18.89& 40.36 \\
Mondi \cite{liu2022monitored}     & 29.67 & 79.78 & 14.84& 37.88 \\ 
NLSPN \cite{park2020non}          & 26.70 & 79.12 & 12.70& 33.88 \\
\hline 
CostDCNet \cite{kam2022costdcnet} & 25.84 & 76.28 & 12.19& 32.13  \\
CluDe (ours)                             & 25.74 & 72.34 & 12.76& 33.92  \\
\hline
\end{tabular}
\label{tab_r}
\end{table}

\begin{figure*}[t]
\centering
\includegraphics[width=0.9\linewidth]{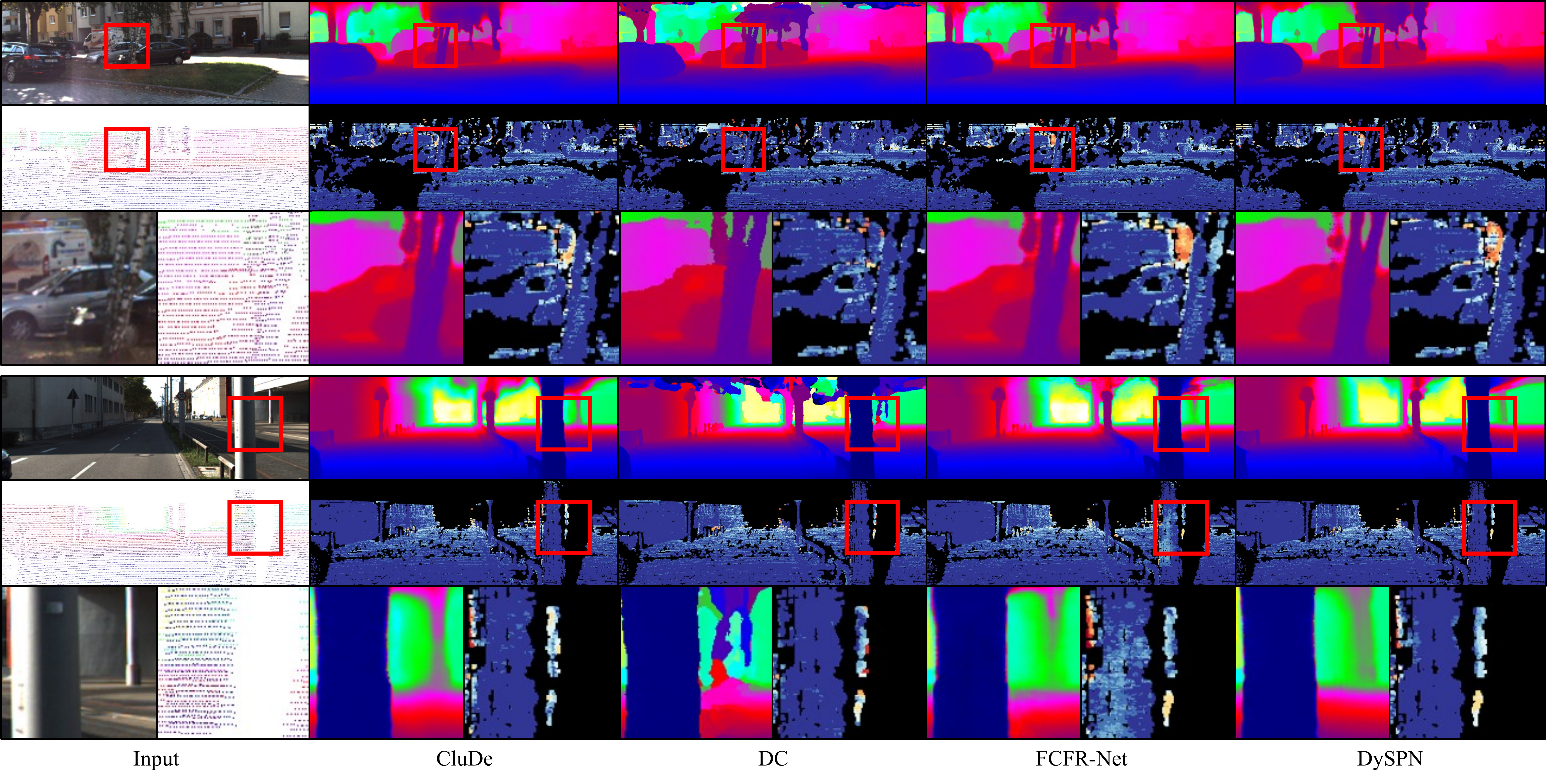}
\caption{Qualitative results on KITTI testing set. For each method, we show the depth result (\textit{top}) and error map (\textit{bottom}). In addition, we zoom in the box region where the boundaries of our method are clearer.}
\label{fig_kitti}
\end{figure*}

In Fig. \ref{fig_kitti}, we present visualizations on the KITTI dataset, which clearly demonstrate that the results obtained by DC contain a significant amount of noise, while CluDe accurately predicts dense depth maps. Even in the boxed regions where the sparse depth map contains noise and holes, CluDe successfully reconstructs the underlying environmental structure.

{\setlength{\parindent}{0cm}\textbf{Comparison with state-of-the-art methods}.} We perform a comprehensive comparison of CluDe with other state-of-the-art methods, including both regression-based and classification-based methods. TABLE \ref{tab_r} shows that, on the KITTI dataset, classification-based methods generally exhibit inferior performance compared to regression-based methods, particularly SPN-based methods like DySPN \cite{lin2022dynamic}. However, classification-based methods offer advantages such as faster convergence and clearer boundaries \cite{fu2018deep,li2022binsformer,imran2019depth}. In comparison to these regression-based methods, CluDe achieves comparable performance to NLSPN in losses of RMSE and MAE. It is only 0.6\% worse than PENet in losses of RMSE, but outperforms PENet by 4.8\% in losses of MAE. Furthermore, CluDe exhibits superiority over some regression-based methods, such as DeepLiDAR \cite{Qiu_2019_CVPR}, and achieves competitive performance compared to GuideFormer \cite{rho2022guideformer} and RigNet \cite{yan2021rignet}. Additionally, CluDe outperforms most regression-based methods in losses of MAE, including DGDF-A \cite{10173567}. On the VOID dataset, CluDe achieves state-of-the-art performance, surpassing other state-of-the-art methods. In Fig. \ref{fig_kitti}, it is evident that CluDe successfully preserves the clear structure when compared to FCFR-Net \cite{liu2021fcfr} and DySPN.

\subsection{Discussion}
\label{dis}
{\setlength{\parindent}{0cm}\textbf{Generalization of our method}.}
We analysis the generalization of our method by comparing it with NLSPN, PENet, ACMNet, and a classification-based method. 
All these methods are trained on the KITTI dataset and subsequently evaluated on the KITTI, Virtual KITTI2 (VKITTI2) \cite{cabon2020vkitti2}, Argoverse \cite{Argoverse}, DIODE \cite{diode_dataset}, and VOID validation sets. As shown in Fig. \ref{fig_dis}, the gap between the training set and the evaluation set presents a substantial challenge to the generalization of all these methods.
As shown in TABLE \ref{tab_ge}, our method works well in KITTI, VKITTI2, Argoverse, DIODE (outdoor), and VOID datasets. This suggests the effectiveness and good generalization ability of our method. This may be attributed to the fact that our method can adaptively adjust depth representation during the clustering process. However, our model is hard to generalize DIODE (indoor) dataset, and this is because the depth clusters of outdoor scenes is significantly different from the these of indoor scenes. In conclusion, our model have a good generalization ability on similar scenes (e.g. the outdoor scenes in different cities), but hard to generalize to the domains with large gaps as other methods (e.g. from outdoor scenes to indoor scenes).

\begin{table*}[t]
\centering
\scriptsize
\caption{Comparison of generalization of various methods. $*$ denotes the training dataset. CLS denotes the classification-based method.}
\begin{tabular}{c|cc|cc|cc|cc|cc|cc}
\hline
\multirow{2}{*}{Method} & \multicolumn{2}{c|}{KITTI$^*$} & \multicolumn{2}{c|}{VKITTI2} & \multicolumn{2}{c|}{Argoverse} & \multicolumn{2}{c|}{\begin{tabular}[c]{@{}c@{}}DIODE\\ outdoor\end{tabular}} & \multicolumn{2}{c|}{\begin{tabular}[c]{@{}c@{}}DIODE\\ indoor\end{tabular}} & \multicolumn{2}{c}{VOID} \\ \cline{2-13} 
& \multicolumn{1}{c}{MAE $\downarrow$}              & \multicolumn{1}{c|}{RMSE $\downarrow$}             & \multicolumn{1}{c}{MAE $\downarrow$}               & \multicolumn{1}{c|}{RMSE $\downarrow$}              & \multicolumn{1}{c}{MAE $\downarrow$}                & \multicolumn{1}{c|}{RMSE $\downarrow$}               & \multicolumn{1}{c}{MAE $\downarrow$}              & \multicolumn{1}{c|}{RMSE $\downarrow$}              & \multicolumn{1}{c}{MAE $\downarrow$}              & \multicolumn{1}{c|}{RMSE $\downarrow$}             & \multicolumn{1}{c}{MAE $\downarrow$}                   & \multicolumn{1}{c}{RMSE $\downarrow$}                  \\
\hline
NLSPN                   & 205.38                               & 802.84                                & 2618.37                               & 7079.74                                & 4230.72                                & 4539.75                                 & 2741.90                              & 6763.98                                & \textbf{791.85}                               & 1861.24                               & \textbf{4730.67}                                   & \textbf{5366.53}                                   \\
PENet                   & 209.00                               & \textbf{757.20}                                & 3271.73                               & 7683.19                                & \multicolumn{1}{c}{-}                  & \multicolumn{1}{c|}{-}                  & 4081.17                              & 7797.93                                & 1348.39                              & \textbf{1787.86}                               & 5705.16                                   & 6287.20                                   \\
ACMNet                  & 212.73                               & 774.23                                & 7589.91                               & 12335.21                               & 8746.54                                & 9148.24                                 & 2689.84                              & 6339.32                                & 1740.77                              & 2560.70                               & 8264.66                                   & 9906.18                                   \\
CLS  & 204.52 & 770.65 & 2845.08  & 7400.33 & 4473.61  & 4775.10 & 2864.99  & 6633.75 & 2409.19 & 3260.87 &5108.26 & 5459.98  \\
CluDe (ours)            & \textbf{199.80}                               & 766.60                                & \textbf{2595.18}                               & \textbf{6806.53}                                & \textbf{2945.13}                                & \textbf{3276.33}                                 & \textbf{2631.81}                              & \textbf{3245.22}                                & 2743.82                              & 3290.79                               & 4883.16                                   & 5731.93         \\
\hline                      
\end{tabular}
\label{tab_ge}
\end{table*}

{\setlength{\parindent}{0cm}\textbf{Efficiency of our method}.} We report the accuracy, latency and parameters to analyze the efficiency in TABLE \ref{tab_eff}. Firstly, it is observed that most methods face challenges in achieving a satisfactory trade-off between accuracy and efficiency, while the parameters of most methods are acceptable.

\begin{table*}[t]
\centering
\caption{Efficiency comparison of our method and related methods on the KITTI test set.}
\begin{tabular}{c|cccc|cc}
\hline
Method & DC \cite{imran2019depth}  & PADNet \cite{peng2022pixelwise} & NLSPN \cite{park2020non} & GuideFormer \cite{rho2022guideformer} & CluDe (ours) &  CluDe$^{\dagger}$ (ours)\\
\hline
RMSE $\downarrow$& 965.87 & 746.19 & 741.68 & 721.48 & 734.59 & 742.26 \\
MAE $\downarrow$ & 215.75 & 197.99 & 199.59 & 207.76 & 200.48 & 197.91  \\
Latency $\downarrow$& 0.15   &  0.03  & 0.22   & 0.22   & 0.14 & 0.14 \\
Param. $\downarrow$ & 27 &  10   &  26   & 33    & 22    & 22 \\
\hline
\end{tabular}
\label{tab_eff}
\end{table*}

For the classification-based method DC, our proposed method outperforms it across all metrics.
Similar to our method, GuideFormer utilizes a transformer structure to deeply interact with multimodal features. In comparison, our CluDe exhibits a 3.63\% improvement in MAE, coupled with a 1.78\% decrease in RMSE. Furthermore, our method achieves improved efficiency, as the transformer inherently supports a clustering-based framework through the attention mechanism, allowing for fewer transformer layers.
NLSPN introduces iterative processing to refine the result. Despite using more iterations than our method, we achieve a 0.97\% improvement in RMSE with faster speed.
PADNet focuses on building a lightweight network using existing efficient frameworks. Although CluDe exhibits lower results in MAE and efficiency compared to PADNet, it showcases a 1.58\% improvement in RMSE. Additionally, CluDe$^{\dagger}$ outperforms PADNet in both RMSE and MAE.

{\setlength{\parindent}{0cm}\textbf{Limitation of our method}.}
Our method based on clustering presents two main limitations. Firstly, our method faces challenges in effectively learning representations for tail categories like most methods. In TABLE \ref{tab_s}, if the depth extends beyond 65m, our method shows inferior performance compared to some classification-based and regression-based methods at specific depth intervals. This limitation arises from challenge in learning depth representations that cover the entire depth range due to the long-tail distribution.

Secondly, we dedicate effort to constructing an effective clustering-based framework that learns pixel depth representations to address depth smearing. Nevertheless, achieving a satisfactory trade-off between accuracy and latency remains a challenge for our method. Although our efficiency surpasses that of NLSPN and GuideFormer, it falls short compared to other methods known for their efficiency, such as PADNet and LRRU \cite{Wang_2023_ICCV}.

\section{CONCLUSION}
In this paper, we discover that pre-defined depth categories cannot represent the continuous depth values present in real depth distribution, leading to depth smearing in the conventional classification-based methods. We revisit depth completion from a clustering perspective and propose a new framework called CluDe, which goes beyond previous methods to learn pixel-wise continuous depth representation from various real depth distributions. Extensive experiments demonstrate that our proposed CluDe can significantly improve the quality of depth results and effectively alleviate the problem of depth smearing. Most importantly, our method achieves state-of-the-art performance on both outdoor and indoor scenes in classification-based depth completion methods. In the future, we plan to address two main challenges: addressing the long-tail distribution issue to improve depth results in tail categories and enhancing the efficiency of our framework.

{\small
	\bibliographystyle{ieee_fullname}
	\bibliography{ref}

\begin{thebibliography}{10}\itemsep=-1pt

\bibitem{cabon2020vkitti2}
Yohann Cabon, Naila Murray, and Martin Humenberger.
\newblock Virtual kitti 2, 2020.

\bibitem{Argoverse}
Ming-Fang Chang, John~W Lambert, Patsorn Sangkloy, Jagjeet Singh, Slawomir Bak,
  Andrew Hartnett, De Wang, Peter Carr, Simon Lucey, Deva Ramanan, and James
  Hays.
\newblock Argoverse: 3d tracking and forecasting with rich maps.
\newblock In {\em Proc. IEEE Conf. Comput. Vis. Pattern Recognit.}, 2019.

\bibitem{chen2019over}
Chuangrong Chen, Xiaozhi Chen, and Hui Cheng.
\newblock On the over-smoothing problem of cnn based disparity estimation.
\newblock In {\em Proc. IEEE Int. Conf. Comput. Vis}, pages 8997--9005, 2019.

\bibitem{10064324}
Guancheng Chen, Junli Lin, and Huabiao Qin.
\newblock Uamd-net: A unified adaptive multimodal neural network for dense
  depth completion.
\newblock {\em IEEE Trans. Circuits Syst. Video Technol.}, pages 1--1, 2023.

\bibitem{cheng2018depth}
Xinjing Cheng, Peng Wang, and Ruigang Yang.
\newblock Depth estimation via affinity learned with convolutional spatial
  propagation network.
\newblock In {\em Proc. Eur. Conf. Comput. Vis.}, pages 103--119, 2018.

\bibitem{dosovitskiy2021an}
Alexey Dosovitskiy, Lucas Beyer, Alexander Kolesnikov, Dirk Weissenborn,
  Xiaohua Zhai, Thomas Unterthiner, Mostafa Dehghani, Matthias Minderer, Georg
  Heigold, Sylvain Gelly, Jakob Uszkoreit, and Neil Houlsby.
\newblock An image is worth 16x16 words: Transformers for image recognition at
  scale.
\newblock In {\em Proc. Int. Conf. Learn. Represent.}, 2021.

\bibitem{9811556}
Wenchao du, Hu Chen, Hongyu Yang, and Yi Zhang.
\newblock Depth completion using geometry-aware embedding.
\newblock In {\em ICRA}, pages 8680--8686, 2022.

\bibitem{fang2020jgr}
Linpu Fang, Xingyan Liu, Li Liu, Hang Xu, and Wenxiong Kang.
\newblock Jgr-p2o: Joint graph reasoning based pixel-to-offset prediction
  network for 3d hand pose estimation from a single depth image.
\newblock In {\em Proc. Eur. Conf. Comput. Vis.}, pages 120--137, 2020.

\bibitem{fu2018deep}
Huan Fu, Mingming Gong, Chaohui Wang, Kayhan Batmanghelich, and Dacheng Tao.
\newblock Deep ordinal regression network for monocular depth estimation.
\newblock In {\em Proc. IEEE Conf. Comput. Vis. Pattern Recognit.}, pages
  2002--2011, 2018.

\bibitem{garg2020wasserstein}
Divyansh Garg, Yan Wang, Bharath Hariharan, Mark Campbell, Kilian~Q Weinberger,
  and Wei-Lun Chao.
\newblock Wasserstein distances for stereo disparity estimation.
\newblock {\em Proc. Adv. Neural Inf. Process. Syst.}, 33:22517--22529, 2020.

\bibitem{geiger2012we}
Andreas Geiger, Philip Lenz, and Raquel Urtasun.
\newblock Are we ready for autonomous driving? the kitti vision benchmark
  suite.
\newblock In {\em Proc. IEEE Conf. Comput. Vis. Pattern Recognit.}, pages
  3354--3361, 2012.

\bibitem{hu2020PENet}
Mu Hu, Shuling Wang, Bin Li, Shiyu Ning, Li Fan, and Xiaojin Gong.
\newblock Penet: Towards precise and efficient image guided depth completion.
\newblock 2021.

\bibitem{huang2021alignseg}
Zilong Huang, Yunchao Wei, Xinggang Wang, Wenyu Liu, Thomas~S Huang, and
  Humphrey Shi.
\newblock Alignseg: Feature-aligned segmentation networks.
\newblock {\em IEEE Trans. Pattern Anal. Mach. Intell.}, 44(1):550--557, 2021.

\bibitem{imran2019depth}
Saif Imran, Yunfei Long, Xiaoming Liu, and Daniel Morris.
\newblock Depth coefficients for depth completion.
\newblock In {\em Proc. IEEE Conf. Comput. Vis. Pattern Recognit.}, pages
  12438--12447, 2019.

\bibitem{kam2022costdcnet}
Jaewon Kam, Jungeon Kim, Soongjin Kim, Jaesik Park, and Seungyong Lee.
\newblock Costdcnet: Cost volume based depth completion for a single rgb-d
  image.
\newblock In {\em Proc. Eur. Conf. Comput. Vis.}, pages 257--274, 2022.

\bibitem{kolodner1992introduction}
Janet~L Kolodner.
\newblock An introduction to case-based reasoning.
\newblock {\em Artificial intelligence review}, 6(1):3--34, 1992.

\bibitem{lee2021depth}
Byeong-Uk Lee, Kyunghyun Lee, and In~So Kweon.
\newblock Depth completion using plane-residual representation.
\newblock In {\em Proc. IEEE Conf. Comput. Vis. Pattern Recognit.}, pages
  13916--13925, 2021.

\bibitem{li2022binsformer}
Zhenyu Li, Xuyang Wang, Xianming Liu, and Junjun Jiang.
\newblock Binsformer: Revisiting adaptive bins for monocular depth estimation,
  2022.

\bibitem{lin2022dynamic}
Yuankai Lin, Tao Cheng, Qi Zhong, Wending Zhou, and Hua Yang.
\newblock Dynamic spatial propagation network for depth completion, 2022.

\bibitem{10284921}
Yuankai Lin, Hua Yang, Tao Cheng, Wending Zhou, and Zhouping Yin.
\newblock Dyspn: Learning dynamic affinity for image-guided depth completion.
\newblock {\em IEEE Trans. Circuits Syst. Video Technol.}, pages 1--1, 2023.

\bibitem{liu2021fcfr}
Lina Liu, Xibin Song, Xiaoyang Lyu, Junwei Diao, Mengmeng Wang, Yong Liu, and
  Liangjun Zhang.
\newblock Fcfr-net: Feature fusion based coarse- to-fine residual learning for
  depth completion.
\newblock In {\em Proc. AAAI Conf. Artif. Intell.}, volume~35, pages
  2136--2144, 2021.

\bibitem{liu2017learning}
Sifei Liu, Shalini De~Mello, Jinwei Gu, Guangyu Zhong, Ming-Hsuan Yang, and Jan
  Kautz.
\newblock Learning affinity via spatial propagation networks.
\newblock {\em Proc. Adv. Neural Inf. Process. Syst.}, 30, 2017.

\bibitem{liu2022monitored}
Tian~Yu Liu, Parth Agrawal, Allison Chen, Byung-Woo Hong, and Alex Wong.
\newblock Monitored distillation for positive congruent depth completion.
\newblock In {\em Proc. Eur. Conf. Comput. Vis.}, pages 35--53, 2022.

\bibitem{liu2022graphcspn}
Xin Liu, Xiaofei Shao, Bo Wang, Yali Li, and Shengjin Wang.
\newblock Graphcspn: Geometry-aware depth completion via dynamic gcns.
\newblock In {\em Proc. Eur. Conf. Comput. Vis.}, pages 90--107, 2022.

\bibitem{liu2021swin}
Ze Liu, Yutong Lin, Yue Cao, Han Hu, Yixuan Wei, Zheng Zhang, Stephen Lin, and
  Baining Guo.
\newblock Swin transformer: Hierarchical vision transformer using shifted
  windows.
\newblock In {\em Proc. IEEE Int. Conf. Comput. Vis.}, pages 10012--10022,
  2021.

\bibitem{long2015fully}
Jonathan Long, Evan Shelhamer, and Trevor Darrell.
\newblock Fully convolutional networks for semantic segmentation.
\newblock In {\em Proc. IEEE Conf. Comput. Vis. Pattern Recognit.}, pages
  3431--3440, 2015.

\bibitem{ma2018sparse}
Fangchang Ma and Sertac Karaman.
\newblock Sparse-to-dense: Depth prediction from sparse depth samples and a
  single image.
\newblock In {\em ICRA}, pages 4796--4803, 2018.

\bibitem{papandreou2017towards}
George Papandreou, Tyler Zhu, Nori Kanazawa, Alexander Toshev, Jonathan
  Tompson, Chris Bregler, and Kevin Murphy.
\newblock Towards accurate multi-person pose estimation in the wild.
\newblock In {\em Proc. IEEE Conf. Comput. Vis. Pattern Recognit.}, pages
  4903--4911, 2017.

\bibitem{park2020non}
Jinsun Park, Kyungdon Joo, Zhe Hu, Chi-Kuei Liu, and In So~Kweon.
\newblock Non-local spatial propagation network for depth completion.
\newblock In {\em Proc. Eur. Conf. Comput. Vis.}, pages 120--136, 2020.

\bibitem{peng2022pixelwise}
Rui Peng, Tao Zhang, Bing Li, and Yitong Wang.
\newblock Pixelwise adaptive discretization with uncertainty sampling for depth
  completion.
\newblock In {\em ACM MM}, pages 3926--3935, 2022.

\bibitem{Qiu_2019_CVPR}
Jiaxiong Qiu, Zhaopeng Cui, Yinda Zhang, Xingdi Zhang, Shuaicheng Liu, Bing
  Zeng, and Marc Pollefeys.
\newblock Deeplidar: Deep surface normal guided depth prediction for outdoor
  scene from sparse lidar data and single color image.
\newblock In {\em Proc. IEEE Conf. Comput. Vis. Pattern Recognit.}, June 2019.

\bibitem{rho2022guideformer}
Kyeongha Rho, Jinsung Ha, and Youngjung Kim.
\newblock Guideformer: Transformers for image guided depth completion.
\newblock In {\em Proc. IEEE Conf. Comput. Vis. Pattern Recognit.}, pages
  6250--6259, 2022.

\bibitem{shen2021cfnet}
Zhelun Shen, Yuchao Dai, and Zhibo Rao.
\newblock Cfnet: Cascade and fused cost volume for robust stereo matching.
\newblock In {\em Proc. IEEE Conf. Comput. Vis. Pattern Recognit.}, pages
  13906--13915, 2021.

\bibitem{9286883}
Jie Tang, Fei-Peng Tian, Wei Feng, Jian Li, and Ping Tan.
\newblock Learning guided convolutional network for depth completion.
\newblock {\em IEEE Trans. Image Process.}, 30:1116--1129, 2021.

\bibitem{diode_dataset}
Igor Vasiljevic, Nick Kolkin, Shanyi Zhang, Ruotian Luo, Haochen Wang,
  Falcon~Z. Dai, Andrea~F. Daniele, Mohammadreza Mostajabi, Steven Basart,
  Matthew~R. Walter, and Gregory Shakhnarovich.
\newblock {DIODE}: {A} {D}ense {I}ndoor and {O}utdoor {DE}pth {D}ataset.
\newblock {\em CoRR}, abs/1908.00463, 2019.

\bibitem{vaswani2017attention}
Ashish Vaswani, Noam Shazeer, Niki Parmar, Jakob Uszkoreit, Llion Jones,
  Aidan~N Gomez, {\L}ukasz Kaiser, and Illia Polosukhin.
\newblock Attention is all you need.
\newblock {\em Proc. Adv. Neural Inf. Process. Syst.}, 30, 2017.

\bibitem{wang2022visual}
Wenguan Wang, Cheng Han, Tianfei Zhou, and Dongfang Liu.
\newblock Visual recognition with deep nearest centroids.
\newblock {\em arXiv preprint arXiv:2209.07383}, 2022.

\bibitem{wang2021pyramid}
Wenhai Wang, Enze Xie, Xiang Li, Deng-Ping Fan, Kaitao Song, Ding Liang, Tong
  Lu, Ping Luo, and Ling Shao.
\newblock Pyramid vision transformer: A versatile backbone for dense prediction
  without convolutions.
\newblock In {\em Proc. IEEE Int. Conf. Comput. Vis.}, pages 568--578, 2021.

\bibitem{Wang_2023_ICCV}
Yufei Wang, Bo Li, Ge Zhang, Qi Liu, Tao Gao, and Yuchao Dai.
\newblock Lrru: Long-short range recurrent updating networks for depth
  completion.
\newblock In {\em Proc. IEEE Int. Conf. Comput. Vis.}, pages 9422--9432,
  October 2023.

\bibitem{10173567}
Yufei Wang, Yuxin Mao, Qi Liu, and Yuchao Dai.
\newblock Decomposed guided dynamic filters for efficient rgb-guided depth
  completion.
\newblock {\em IEEE Trans. Circuits Syst. Video Technol.}, pages 1--1, 2023.

\bibitem{weng2021stage}
Xi Weng, Yan Yan, Si Chen, Jing-Hao Xue, and Hanzi Wang.
\newblock Stage-aware feature alignment network for real-time semantic
  segmentation of street scenes.
\newblock {\em IEEE Trans. Circuits Syst. Video Technol.}, 32(7):4444--4459,
  2021.

\bibitem{wong2020unsupervised}
Alex Wong, Xiaohan Fei, Stephanie Tsuei, and Stefano Soatto.
\newblock Unsupervised depth completion from visual inertial odometry.
\newblock {\em IEEE Robot. Autom. Lett.}, 5(2):1899--1906, 2020.

\bibitem{xiong2019a2j}
Fu Xiong, Boshen Zhang, Yang Xiao, Zhiguo Cao, Taidong Yu, Joey~Tianyi Zhou,
  and Junsong Yuan.
\newblock A2j: Anchor-to-joint regression network for 3d articulated pose
  estimation from a single depth image.
\newblock In {\em Proc. IEEE Int. Conf. Comput. Vis.}, pages 793--802, 2019.

\bibitem{xu2022groupvit}
Jiarui Xu, Shalini De~Mello, Sifei Liu, Wonmin Byeon, Thomas Breuel, Jan Kautz,
  and Xiaolong Wang.
\newblock Groupvit: Semantic segmentation emerges from text supervision.
\newblock In {\em Proc. IEEE Conf. Comput. Vis. Pattern Recognit.}, pages
  18134--18144, 2022.

\bibitem{Xu_2023_CVPR}
Mengde Xu, Zheng Zhang, Fangyun Wei, Han Hu, and Xiang Bai.
\newblock Side adapter network for open-vocabulary semantic segmentation.
\newblock In {\em Proc. IEEE Conf. Comput. Vis. Pattern Recognit.}, pages
  2945--2954, June 2023.

\bibitem{yan2021rignet}
Zhiqiang Yan, Kun Wang, Xiang Li, Zhenyu Zhang, Jun Li, and Jian Yang.
\newblock Rignet: Repetitive image guided network for depth completion.

\bibitem{10243117}
Linqing Zhao, Wenzhao Zheng, Yueqi Duan, Jie Zhou, and Jiwen Lu.
\newblock Sptr: Structure-preserving transformer for unsupervised indoor depth
  completion.
\newblock {\em IEEE Trans. Circuits Syst. Video Technol.}, pages 1--1, 2023.

\bibitem{zhao2021adaptive}
Shanshan Zhao, Mingming Gong, Huan Fu, and Dacheng Tao.
\newblock Adaptive context-aware multi-modal network for depth completion.
\newblock {\em IEEE Trans. Image Process.}, 30:5264--5276, 2021.

\bibitem{10.1145/3503161.3548381}
Zhengming Zhou and Qiulei Dong.
\newblock Learning occlusion-aware coarse-to-fine depth map for self-supervised
  monocular depth estimation.
\newblock In {\em ACM MM}, page 6386–6395, 2022.

\end{thebibliography}
}

\begin{IEEEbiography}
[{\includegraphics[width=1in,height=1.25in,clip,keepaspectratio]{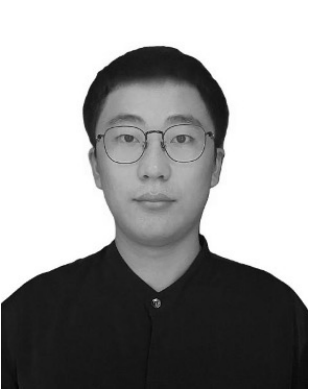}}]{Shenglun Chen}
received the B.S. and M.S. degree in School of Software, Dalian University of Technology in 2017 and 2021.
He is currently a doctoral candidate of the School of Software, Dalian University of Technology, Liaoning, China. His main research interests include depth estimation, stereo matching and SLAM.
\end{IEEEbiography}

\begin{IEEEbiography}
[{\includegraphics[width=1in,height=1.25in,clip,keepaspectratio]{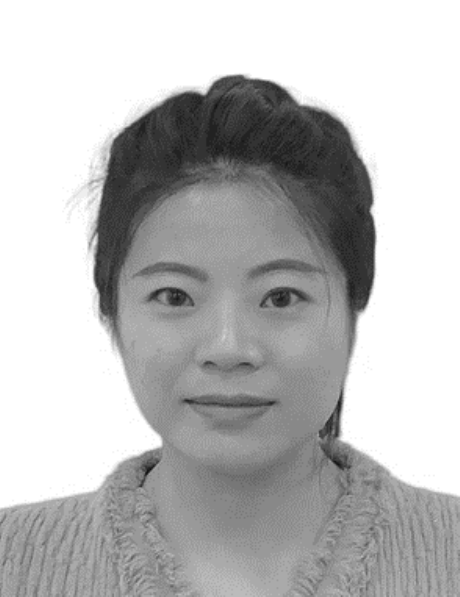}}]{Hong Zhang}
received the B.S. and Ph.D. degree in School of Software, Dalian University of Technology in 2017. She is currently a doctoral candidate of the School of Software, Dalian University of Technology, Liaoning, China. Her research interests include 3D visual perception and multi-modal remote sensing image matching.
\end{IEEEbiography}

\begin{IEEEbiography}
[{\includegraphics[width=1in,height=1.25in,clip,keepaspectratio]{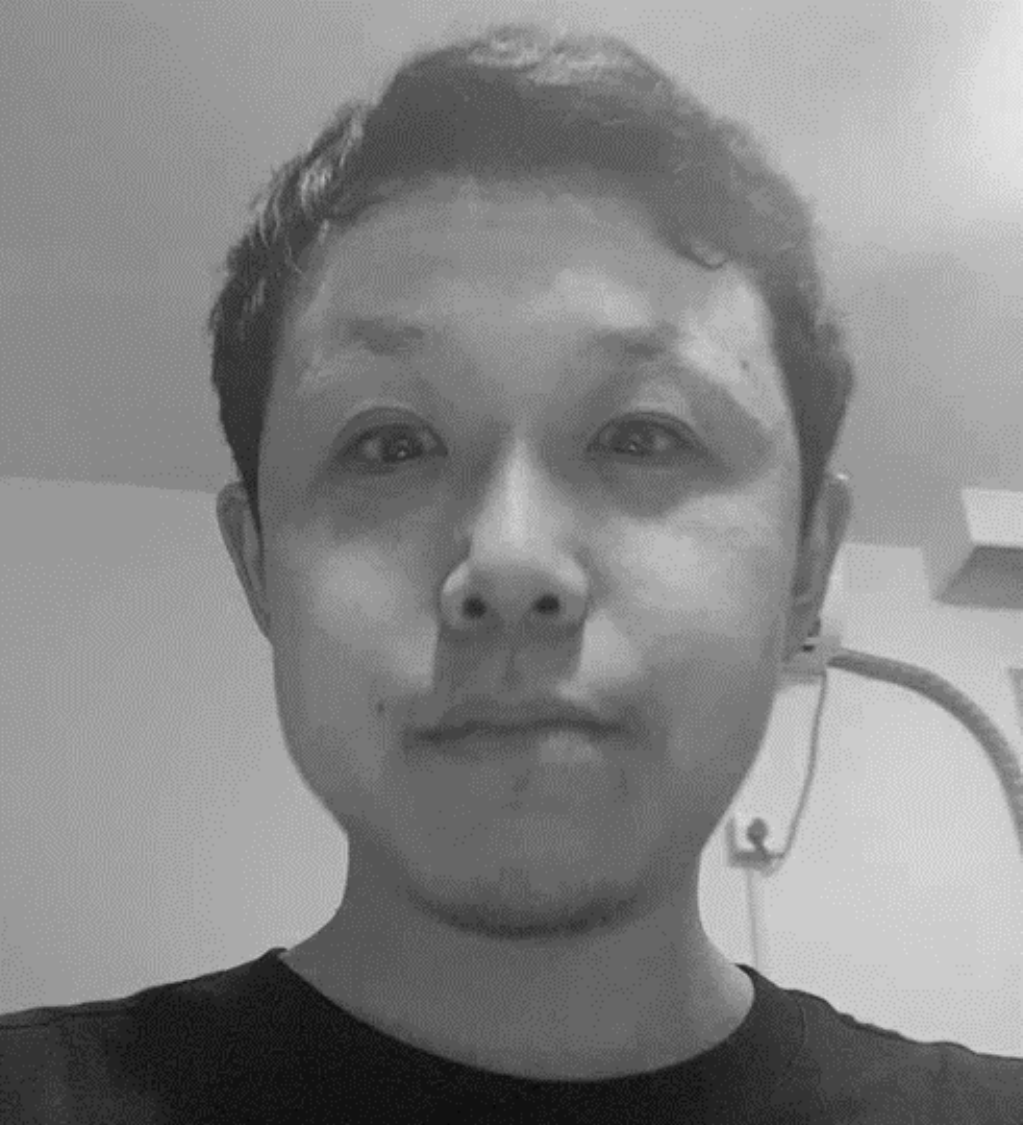}}]{Xinzhu Ma}
received his B.Eng and M.P’s degree from Dalian University of Technology in 2017 and 2019, respectively. After that, He got the Ph.D. degree from the University of Sydney in 2023. He is currently a postdoctoral researcher at the Chinese University of Hong Kong. His research interests include deep learning and computer vision.
\end{IEEEbiography}

\begin{IEEEbiography}
[{\includegraphics[width=1in,height=1.25in,clip,keepaspectratio]{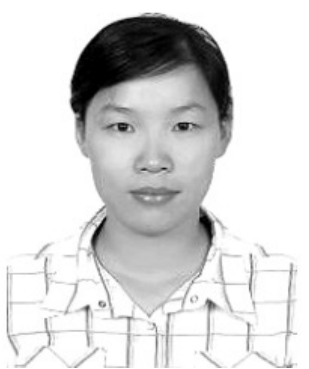}}]{Zhihui Wang}
received the B.S. degree in software engineering from North Eastern University, Shenyang, China, in 2004, and the M.S. degree in software engineering and the Ph.D. degree in software and theory of computer from the Dalian University of Technology, Dalian, China, in 2007 and 2010, respectively. Since November 2011, she has been a Visiting Scholar with the University of Washington. Her current research interests include
information hiding and image compression.
\end{IEEEbiography}

\begin{IEEEbiography}
[{\includegraphics[width=1in,height=1.25in,clip,keepaspectratio]{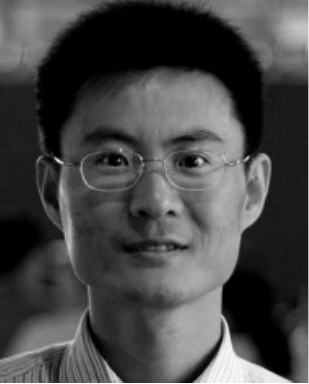}}]{Haojie Li}
received the B.E. degree from Nankai University, Tianjin, in 1996 and the Ph.D. degree from the Institute of Computing Technology,Chinese Academy of Sciences, Beijing, in 2007. From 2010 to 2023 he was a full Professor with the International School of Information Science \& Engineering, Dalian University of Technology.  He is currently a full Professor with the College of Computer Science and Engineering, Shandong University of Science and Technology. His research interests include social media computing and multimedia information retrieval.
\end{IEEEbiography}

\end{document}